\pdfoutput=1

\PassOptionsToPackage{table,xcdraw}{xcolor}
\documentclass[11pt]{article}

\usepackage[final]{acl}

\usepackage{times}
\usepackage{latexsym}
\usepackage{xcolor}
\usepackage{fontawesome} 

\usepackage[T1]{fontenc}

\usepackage[utf8]{inputenc}

\usepackage{microtype}

\usepackage{inconsolata}

\usepackage{graphicx}
\usepackage{booktabs}
\usepackage{dingbat}
\usepackage{amsmath}
\usepackage{algorithm}
\usepackage{algpseudocode}
\usepackage{wrapfig}
\usepackage{multirow}
\usepackage{subcaption}
\usepackage{hyperref}

\title{\textsc{Select}LLM: Query-Aware Efficient Selection Algorithm \\ for Large Language Models}

\author{
    Kaushal Kumar Maurya\thanks{Equal contribution} \\
    \And
    KV Aditya Srivatsa\footnotemark[1] \\
    Mohamed Bin Zayed University of Artificial Intelligence, Abu Dhabi, UAE \\
    \texttt{\{kaushal.maurya, vaibhav.kuchibhotla, ekaterina.kochmar\}@mbzuai.ac.ae} \\
    \And
    Ekaterina Kochmar
}

\begin{document}
\maketitle
\begin{abstract}
Large language models (LLMs) have been widely adopted due to their remarkable performance across various applications, driving the accelerated development of a large number of diverse models. However, these individual LLMs show limitations in generalization and performance on complex tasks due to inherent training biases, model size constraints, and the quality or diversity of pre-training datasets. A promising direction is to efficiently harness the diverse capabilities of LLMs to overcome these individual limitations. To address these limitations, we introduce a novel LLM selection algorithm called \textsc{SelectLLM}, which efficiently directs input queries to the most suitable subset of LLMs from a large pool, ensuring that the selected models collectively provide accurate responses. \textsc{SelectLLM} employs a multi-label classifier and policy based on the classifier's predictions and confidence scores in selecting an optimal, query-aware, and lightweight subset of LLMs. Our findings indicate that the proposed model outperforms existing ensemble-based baselines and achieves competitive performance with similarly sized top-performing LLMs while maintaining efficiency. Specifically, it achieves a huge reduction in inference latency on two challenging reasoning benchmarks: 13\% on {\tt GSM8K} and 70\% on {\tt MMLU}, compared to the top-performing baseline. Also, we establish a theoretical upper bound by an Oracle with LLMs and perform an in-depth linguistic analysis to understand the performance gap between the Oracle and \textsc{SelectLLM}.

\href{https://github.com/kaushal0494/SelectLLM/}{\faGithub\ \texttt{github.com/kaushal0494/SelectLLM}}

\end{abstract}


\begin{figure}
    \centering
    \includegraphics[width=1\linewidth]{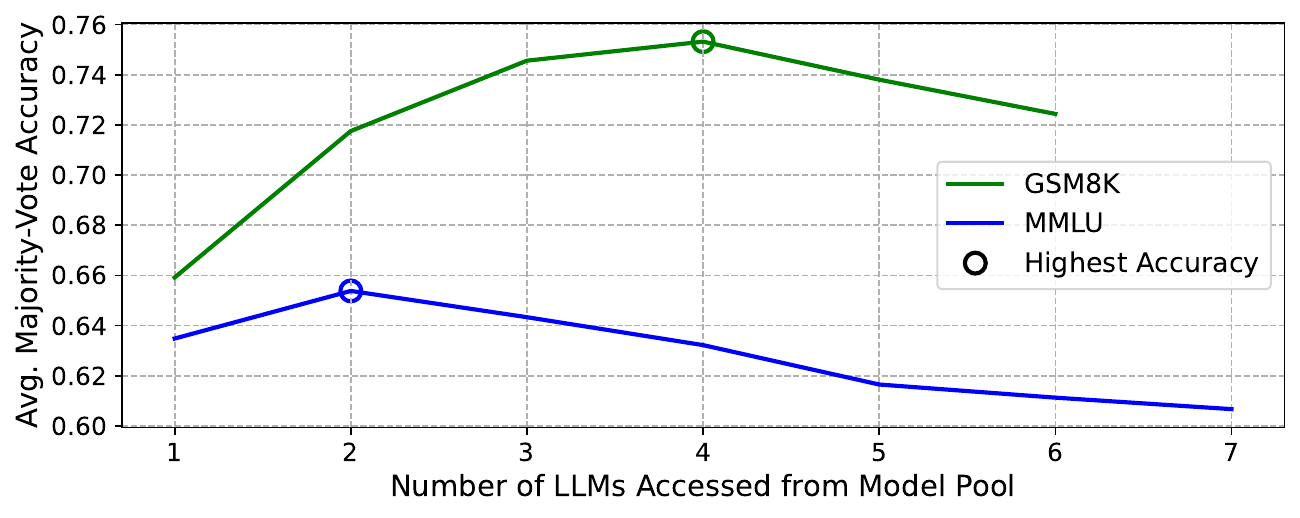}
    \caption{Accuracy (using majority voting) versus the number of LLMs plot for the {\tt GSM8K} and {\tt MMLU} test sets. Models are added in descending order of their performance on their corresponding training sets.}
    \label{fig:pool_vs_acc}
    \vspace{-0.3 cm}
\end{figure}

\section{Introduction}
\label{sec:intro}

In recent years, large language models (LLMs) have demonstrated remarkable capabilities in solving a wide range of core NLP tasks \citep{bommasani2021opportunities, chang2023survey}. Despite these advances, existing LLMs still struggle with complex tasks such as factually-grounded reasoning and planning \citep{wei2022chain, kojima2022large, minaee2024large}. Moreover, the wide range of LLMs available seem to \textbf{exhibit diverse capabilities} \citep{jiang-etal-2023-llm}, resulting in no single (especially open-source) LLM being effective across all benchmarks and datasets. Aligning with recent research trends \cite{deepseekai2025deepseekr1incentivizingreasoningcapability, mirzadeh2024gsm}, \textit{we focus on natural language understanding tasks grounded in reasoning, specifically those with discrete output values, and do not consider general natural language generation tasks.}

Although newer and more powerful models are constantly introduced, an alternative and cost-effective approach involves harnessing the diverse capabilities of existing models to improve the overall response quality using ensembling \citep{wang2022self, wang2023fusing, li2024more} and collaborative frameworks \citep{wu2023autogen, li2023camel}. However, these approaches often require access to the responses from all models in the pool to choose the optimal response(s), which greatly increases the overall computational cost for such ensembles.

The individual LLMs exhibit diverse capabilities, i.e., \textit{not all models may be suited for all kinds of tasks}. Fig.~\ref{fig:pool_vs_acc} reports the accuracy of a LLM model pool (spanning up to 7 diverse LLMs) on two challenging reasoning benchmarks -- {\tt GSM8K} \citep{cobbe2021training} and {\tt MMLU} \citep{hendryckstest2021}. As the plot demonstrates, utilizing more LLMs initially improves performance, which is supported by previous research towards employing more LLMs and more responses per model \citep{li2024more}. However, note that using more (or even all) models in the pool does not necessarily result in the best scores overall. Thus, \textit{selectively abstaining} from querying unsuitable LLMs for a given task may help \textbf{improve the overall response quality} of such ensembles. Additionally, such an approach would \textbf{implicitly save computational resources} by accessing fewer models per query.

In this paper, we propose the novel \textsc{SelectLLM} algorithm to explore this idea. Our approach first employs a \textit{multi-label classifier} to learn the LLM-specific capabilities using a dataset of diverse queries. When running inference for a unseen query, this knowledge is utilized to predict confidence scores for each LLM in the model pool, reflecting \textit{their likelihood of successfully solving the task}. Next, we develop various \textit{selection policies} to determine the optimal subset of LLMs for each query based on these confidence scores and predictions. Additionally, we establish a theoretical Oracle model's upper bound that can be achieved collectively by all LLMs and perform a \textit{qualitative and quantitative linguistic analysis} of the inputs to understand the performance gap between the Oracle and \textsc{SelectLLM}. The contributions of our work are as follows: \vspace{-0.5em}

\begin{itemize}
    \item We introduce the novel \textsc{SelectLLM} algorithm, which is based on a multi-label classifier and an optimal confidence-based policy. This approach efficiently navigates input queries to the ideal subset of LLMs from a larger pool to improve response quality and simultaneously reduce computational costs.\vspace{-0.5em}

    \item The efficacy of the proposed \textsc{SelectLLM} algorithm is evaluated on two challenging reasoning benchmarks. We report an improvement of 1.90 points on {\tt GSM8K} and 4.89 points in terms of accuracy on {\tt MMLU} compared to the existing strong ensemble-based baselines and competitive performance with a model, which consists of similarly sized top-performing LLM subsets. Additionally, we observe significantly lower inference latencies, with reductions of 13\% for {\tt GSM8K} and 70\% for {\tt MMLU}, compared to the top-performing baseline.\vspace{-0.5em}
    
    \item We present a theoretical upper bound established by an Oracle model, representing the maximum performance achievable collectively by all LLMs in our pool. Furthermore, we conduct a linguistic feature-based analysis of the inputs to understand the gains achieved by \textsc{SelectLLM} and the performance gap relative to the Oracle model.\vspace{-0.5em}

\end{itemize}

\begin{figure*}[!t]
    \centering
    \begin{minipage}[t]{0.59\textwidth}
        \centering
        \includegraphics[width=\textwidth]{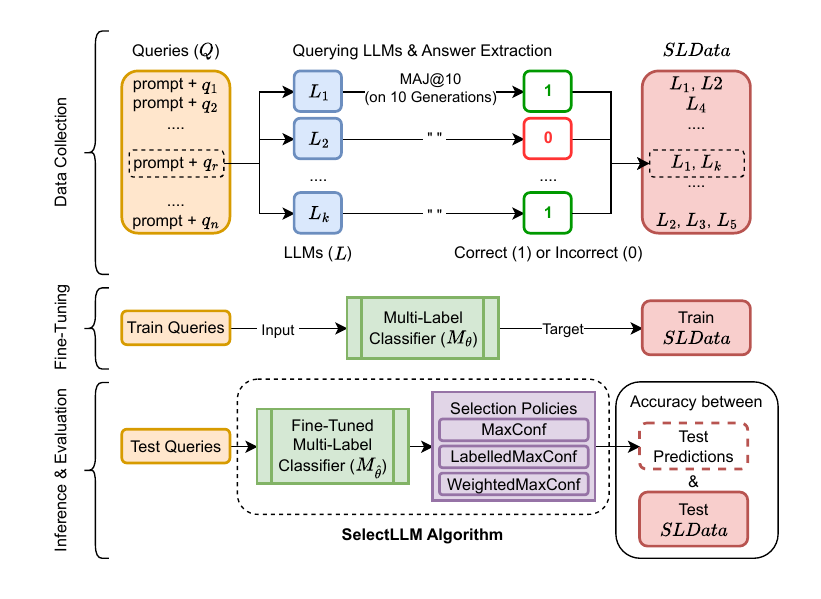}
        \vspace{-1.2cm}
        \caption{Overview of the proposed workflow.}
        \label{fig:workflow}
    \end{minipage}%
    \hfill
    \begin{minipage}[t]{0.40\textwidth}
        \vspace{-5.5cm}
        \centering
        \begin{table}[H]
            \centering
            \resizebox{0.70\linewidth}{!}{%
            \begin{tabular}{l|c|c|c|c}
            \hline \hline
            \textbf{LLM} & \multicolumn{2}{c|}{\textbf{GSM8K}} & \multicolumn{2}{c}{\textbf{MMLU}} \\ \cline{2-5}
                         & \textbf{Train} & \textbf{Test} & \textbf{Train} & \textbf{Test} \\ \hline
            \texttt{llama2-7b-lm} & 22.21 & 24.01 & 46.85 & 48.10 \\
            \texttt{gemma-7b-lm} & 73.49 & 71.27 & 66.10 & 63.73 \\
            \texttt{mistral-7b-lm} & 59.33 & 60.50 & 61.69 & 61.57\\
            \texttt{metamath-7b-lm} & 92.35 & 67.25 & 42.16  & 41.46\\ \hline
            \texttt{*gemma-7b-it} & 41.62 & 42.23 & 50.61 & 50.72\\
            \texttt{llama2-13b-chat} & 50.37 & 49.20 & 54.40 & 52.94 \\
            \texttt{*mistral-7b-it} & 55.92 & 56.71 & 53.39 & 53.92\\ \hline \hline
            \end{tabular}%

            }
            \caption{\small Accuracy with majority voting ({\sc maj@M}) for considered LLMs on {\tt GSM8K} and {\tt MMLU} datasets with train and test splits. All scores were calculated over M response generations for each LLM. Here we use M=10 inspired by \citet{li2024more}.}
            \label{tab:llm_performance}
        \end{table}
    \end{minipage}
    \vspace{-0.4cm}
\end{figure*}


\section{Related Work}
\label{sec:relwork}

\paragraph{LLM Diversity and Capabilities:} LLMs exhibit emergent capabilities, enabling them to perform beyond their explicit training objectives, with diversity in training data fostering broad domain expertise~\citep{bommasani2021opportunities, minaee2024large}. This diversity extends to architectural variations, multilingual proficiency, and adaptability across domains, making them highly effective for a wide range of tasks. Additionally, LLMs demonstrate strong generalization and reasoning abilities, enabling them to tackle complex challenges such as question answering, summarization, and classification~\citep{hendryckstest2021, cobbe2021training, joshi-etal-2017-triviaqa, tam-etal-2023-evaluating, zhang2023sentiment}.  Despite their strengths, no single open-source LLM consistently outperforms others across benchmarks~\citep{jiang-etal-2023-llm}, highlighting the necessity of ensemble methods to leverage diverse model strengths. This work focuses on leveraging the diverse capabilities of LLMs in \textsc{SelectLLM} to address individual LLM limitations.



\paragraph{LLM Ensembling} 
Previous attempts at ensembling and routing LLMs typically fall into three categories: (1) Selecting the best response from multiple LLM generations, as seen in \citet{liu-liu-2021-simcls}, \citet{ravaut-etal-2022-summareranker}, and \citet{jiang-etal-2023-llm}. However, this approach requires querying all LLMs in the model pool for each query during inference, which can be computationally expensive with a large number of LLMs. (2) Minimizing the number of queries to larger LLMs to reduce latency and computational costs, as demonstrated by \citet{shnitzer2023large} and \citet{ding2024hybrid}, who redirect simpler queries to the smallest model capable of handling the task. This minimizes querying costs with minimal performance drop. Routing to the single-best LLM while balancing both accuracy and efficiency has proven to be challenging \citep{srivatsa2024harnessing}. (3) Using multiple LLMs in a multi-agent collaboration setting \cite{tran2025multi}, where specialized LLMs work together to solve complex problems. However, it remains unclear how collaboration functions when non-specialized models are involved. The proposed model lays the foundation for addressing this gap. Moreover, we aim to develop an algorithm that improves response accuracy beyond individual LLMs and their combinations by querying only a subset of LLMs expected to be capable of solving the given query. This, in turn, reduces computational costs and latency by avoiding unnecessary queries to unsuitable LLMs.

\section{Problem Setting}
We propose an ensembling-based LLM inference algorithm -- \textsc{SelectLLM} -- to \textit{efficiently} select the \textit{most suitable} query-aware few LLMs from a large pool of available LLMs. The algorithm harnesses the diverse capabilities of different LLMs and selects a subset of models for the input query, jointly leading to the correct response, and this selection of a small subset leads to the reduction in latency.

Formally, for a given set of input queries \( \mathcal{Q} = \{ q_1, q_2, \dots, q_n \} \) and a pool of LLMs \( \mathcal{L} = \{ l_1, l_2, \dots, l_k \} \), the objective is to learn a \textit{selection model} \( \mathcal{M} \) that selects a subset of LLMs \( \mathcal{L}_s \subseteq \mathcal{L} \), which jointly produce the correct answer for a given input query \( q_i \in \mathcal{Q} \), such that the cumulative latency satisfies \( \text{latency}(\mathcal{L}_s) < \text{latency}(\mathcal{L}) \). In the best-case scenario, a query \( q_i \) is processed by a single LLM, while in the worst-case scenario, \( q_i \) is processed by all LLMs in \( \mathcal{L} \).


 

\section{Methodology}
\label{sec:method}

\subsection{LLM Sampling}
\label{sec:llmsample}

\subsubsection{Selection of Benchmarks and LLMs} 
As discussed in Section \ref{sec:intro}, LLMs often struggle with reasoning tasks. To advance modeling and performance in this domain, we have selected two challenging benchmarks. The \texttt{GSM8K} dataset, introduced by \citet{cobbe2021training}, contains 8,792 grade-school level math word problems (MWPs) in English, focusing on mathematical reasoning. The second benchmark, \texttt{MMLU}, proposed by \citet{hendryckstest2021}, includes 14,572 multiple-choice questions across 57 subjects, assessing multi-domain natural language understanding and reasoning. See Appendix Table \ref{tab:data_stats} for detailed dataset statistics.

We have selected a \textit{diverse} and \textit{sparse} set of LLMs based on \textit{explicit} and \textit{implicit} criteria. The explicit criteria includes performance on benchmarks, training methodologies, model specialization, and modes of operation (chat vs. non-chat), among others. Some of these diverse attributes are presented in the Appendix Table \ref{tab:div_llms}. The implicit criteria include factors such as diverse inference latencies (refer to Table \ref{tab:llm_lat}) and prompting types (i.e., zero-shot vs. few-shot), among others. Further, we consider relatively small open-source LLMs (yet representative of exiting open-source LLM space) because: (i) Experiments with these LLMs are suitable for an academic lab setup, and (ii) This aligns with the research trend towards developing LLMs suitable for small mobile devices \cite{abdin2024phi}. Furthermore, since these selection criteria ensure wide representativeness of the LLM pool, we hypothesize that if the proposed approach works for this pool of LLMs, it should be LLM pool-agnostic, although we leave the proof of this to future work. See Appendix  \ref{sec:pmtsamp} for more details. 


\subsubsection{Data Preparation for the \textsc{SelectLLM} Model: \textsc{SLData}} \label{para:sldata}

In this study, we evaluate the performance of each LLM by generating $M$ responses from it for each input query to ensure reliable and replicable behavior of the proposed model. We employ \textit{Majority Voting} \cite{li2024more} to assess whether a query is correctly answered by the LLM or not. Specifically, \textit{Majority Voting}({\sc maj@M} $\in \{0,1\}$) determines whether the most frequent answer from an LLM matches (using \textit{exact string match}) the gold answer or not. The accuracy with {\sc maj@M} across all input prompts is reported in Table \ref{tab:llm_performance}. In the rest of this paper, we consider only those LLMs for which the viable extracted answers are above 90\% (see more details in Appendix Section \ref{sec:pmtsamp}) to ensure response reliability, resulting in 6 acceptable LLMs for the \texttt{GSM8K} dataset and 7 for the \texttt{MMLU} dataset. We prepare the training dataset for the multi-label classification module of {\sc SelectLLM} (detailed in the next section) by associating each input query with the LLM(s) whose majority vote answer (across $M$ samples) matches the gold answer, i.e., {\sc maj@M} $= 1$. Formally, the target label for a query prompt $q_i \in \mathcal{Q}$ is given by $label\left ( q_i \right ) = \left \{ l\: |\: l \in \mathcal{L}, maj@M\left ( q_i,l \right ) = 1 \right \}$. We denote this dataset as \textsc{SLData} which is separately prepared for both \texttt{GSM8K} and \texttt{MMLU}.

\begin{algorithm*}
\small
\caption{{\sc SelectLLM} Inference Algorithm with {\sc WeightedMaxConf} Policy}
\begin{algorithmic}[1]
\Require Queries $\mathcal{Q}$, LLMs $\mathcal{L}$, {\sc SLData} $(\mathcal{X}, \mathcal{Y}) \in \mathcal{D}$, pre-trained model $\mathcal{M}_{\theta}$ with parameters $\theta$
\State Fine-tune the model: $\mathcal{M}_{\hat{\theta}} = \underset{\theta}{\arg\min} \sum_{(x_i, y_i) \in (\mathcal{X}, \mathcal{Y})} \text{Loss}(\mathcal{M}_{\theta}(x_i), y_i)$
\Comment{Fine-tuning $\mathcal{M}_{\theta}$ with $\mathcal{D}$}
\For{each query $q_i$ in $\mathcal{Q}$}
     \State Perform inference: $q_i^{logits}, q_i^{labels} = \mathcal{M}_{\hat{\theta}}(q_i)$
     \Comment{Using fine-tuned model $\mathcal{M}_{\hat{\theta}}$}
     \State Calculate confidence scores: $c_i = \sigma(q_i^{logits})$
     \Comment{$\sigma$ is the sigmoid activation function}
     \State Select top-$s$ confidence scores: $c_i^{(s)} = \max(c_i, s)$ and associated LLMs: $L_s = L(c_i^{(s)})$
     \State Initialize an empty set for answer set $A_i \leftarrow  \phi$
     \For{each $l_j$ in $L_s$}
         \State Generate M responses: $a_i^{M} = l_j(q_i)$
         \Comment{Generate M responses with LLM $l_j$}
        \State Find answer frequency $a_i^f = \{ a_i^k : \text{countof}(a_i^k)/\sqrt{c_i^{(s)j}} \,|\, \text{for } a_i^k \text{ in unique}(a_i^{K})\}$
        \State $A_i \leftarrow A_i  \cup a_i^f$
     \EndFor
     \State Return most frequent answer in $A_i$
\EndFor
\end{algorithmic}
\label{select_algo}
\end{algorithm*}

\subsection{The Proposed \textsc{SelectLLM} Algorithm}
\label{sec:selectllm}

Building on the discussion in Section \ref{sec:intro}, prior research \citep{ding2024hybrid} indicates that easy queries are correctly solved by smaller or general-purpose LLMs, whereas complex queries necessitate the use of specialized or larger LLMs. Conversely, there are rare queries that are incorrectly responded to by large LLMs but correctly answered by smaller LLMs \cite{nezhurina2024alice}. Due to a lack of widely established query-to-LLM mappings, brute-force approaches are typically employed, querying every available LLM to obtain correct answers. As LLMs continue to advance rapidly in large numbers, such approaches become computationally inefficient and sometimes infeasible. One promising approach involves identifying and directing input queries to the most suitable subset of LLMs from a large pool, which jointly respond correctly. This ensures that the query is accurately addressed while maintaining lower latency compared to running inference on all LLMs in the pool.  


Towards this objective, we introduce the query-aware \textsc{SelectLLM} algorithm, designed to select a tailored subset of LLMs, taking into account the nature of the query and enabling them to collaboratively provide correct responses efficiently. The \textsc{SelectLLM} algorithm comprises two primary components: (i) A \textit{multi-label classifier} (\textsc{MLC}) which is fine-tuned with {\sc SLData} dataset, and (ii) A \textit{selection policy}, which utilizes \textsc{MLC}'s prediction  and confidence scores (i.e., the likelihood of an LLM responding correctly to the query) to determine the suitable subset of LLMs.

\paragraph{Multi-label Classifier (MLC)} 
As indicated by prior studies \cite{hu2024slm}, lightweight language models such as BERT, RoBERTa, and T5 exhibit negligible lower query inference latency compared to LLMs. So, we developed a multi-label classifier (MLC) based on these models. The models were fine-tuned using \textsc{SLData} with a multi-label classification objective, incorporating label imbalance techniques \cite{zhang2020towards} for both the \texttt{GSM8K} and \texttt{MMLU} datasets. The results are provided in Table \ref{tab:mlp_results}. Among the tested models, the RoBERTa-based MLC outperformed the others, achieving weighted F1 scores of 0.71 and 0.68 for the \texttt{GSM8K} and \texttt{MMLU} test datasets, respectively. The fine-tuned model predicts a subset of LLMs (i.e., LLM identities) best suited to address the input query and confidence scores \(C\) for each model in the pool.


\paragraph{Confidence-Based Policies} In this section, we discuss how the confidence scores $C$ are utilized to select a suitable subset of LLMs $\mathcal{L}_s$ for each input query. $C$ is defined as $\{ c_1, c_2, \dots, c_i, \dots c_n\}$ where $c_i$ is the confidence scores for the $i^{th}$ input query. Each $c_i$ is further represented as $\{c_i^{l_1}, c_i^{l_2}, \dots c_i^{l_j} \dots c_i^{l_k}\}$ where  $c_i^{l_j}$ is the confidence score of the $j^{th}$ LLM for the $i^{th}$ query. 

The performance of \textsc{SelectLLM} is determined by the selection policy used. For example, a greedy policy that always selects the LLM (or set of LLMs) with the highest confidence may be suboptimal. This is because another LLM (or set of LLMs) in the pool might have higher accuracy but may not be chosen due to slightly lower confidence. Additionally, when there are two subsets with similar confidence and accuracy, it is more efficient to select the subset with lower cumulative latency. Considering these aspects, we propose the following three optimal policies: \vspace{-0.5em}

\begin{enumerate}
    \itemsep0em
    \item \textsc{\textbf{LabelledMaxConf:}} This policy selects the top-$s$ LLMs (\(L_s\)) for an input query \(q_i\) based on two constraints: (i) the LLMs should be present in the MLC predictions, and (ii) Only those LLMs that have confidence scores within the top-$s$ from $c_i$ are considered.\vspace{-0.5em}
    \item \textsc{\textbf{MaxConf:}} This is a more flexible policy than \textsc{LabelledMaxConf} as it only takes into account the second constraint, i.e., it selects the top-$s$ LLMs corresponding to the top-$s$ confidence scores from \(c_i\).\vspace{-0.5em}
   \item \textsc{\textbf{WeightedMaxConf:}} This policy begins by selecting the top-$s$ LLMs based on their high confidence scores, i.e., for a given query $q_i$, we denote the selected LLMs as $L_s^{q_i}$. Subsequently, we modify the frequency of answer values extracted from the responses of each selected LLM, which involves dividing the frequency of each value by the square root of the confidence score associated with the respective LLM. Finally, we collect all response values and their modified frequencies across the selected LLMs (frequencies are added for the same value). The value with the highest frequency after the modification is selected as the final response. The formal steps are presented in Algorithm \ref{select_algo}. Intuitively, dividing by the square root of the confidence score aims to mitigate biased selection of LLM by the policy similar to \cite{wu2024confidence}. This adjustment ensures fairer opportunities for each selected LLM to contribute to the majority voting.  
\end{enumerate}\vspace{-0.5em}

Across all three policies, in case of a conflict where two LLMs have similar confidence, the \textit{light-weight} LLM (i.e., the lower latency) is preferred.
\section{Experimental Setup}
\label{sec:expsetup}

\subsection{Baseline Models}
Based on recent literature, the following baseline models are included for comparison:
\begin{enumerate}
    \itemsep0.1em
    \item \textbf{Oracle:} The maximum performance is assumed under the premise that an Oracle always selects the lowest latency subset of LLMs that generates the correct majority vote answer for each question (if possible; otherwise, the question attempt is marked as incorrect). Empirically, this is obtained by evaluating all subsets of LLMs, i.e., ($2^{k} - 1$), where $k$ is the total number of LLMs. \vspace{-0.3em}
    \item \textbf{Random:} This represents the mean performance of uniformly randomly selecting an LLM subset from all possible ($2^{k} - 1$) subsets for each query. We report mean scores across 1,000 independent runs to avoid biases. \vspace{-0.3em}
   \item \textbf{All LLMs}: This baseline reports the mean accuracy of {\sc maj@(M$\times |L|$)} based on the combined pool of $M$ generations from each LLM and is similar to \citet{li2024more}. \vspace{-0.3em}
   \item \textbf{LLM-Blender} \cite{jiang-etal-2023-llm}: An ensembling framework was developed to utilize the diverse strengths of multiple open-source LLMs. Specifically, it employs \textsc{PairRanker}, which utilizes a cross-attention-based method for pairwise comparison of different LLM responses to determine the superior one. We use the officially released model checkpoint in our setting.\vspace{-0.3em}
   \item \textbf{Top-\textit{s} LLMs:} For this baseline, we consider the responses of the top-s scoring LLMs using a majority-vote strategy. The top-performing models are determined by comparing the overall accuracies of the LLMs.
\end{enumerate}\vspace{-0.3em}

 For \textit{All LLMs}, \textit{LLM-Blender} and \textit{Top-\textit{s} LLMs}, the latency remains constant since they need inference with all LLMs to determine the performance.

\subsection{Evaluation Metrics} 
We evaluate the performance of all models with the \textit{accuracy} ({\tt Acc}) metric using majority voting (see Section \ref{para:sldata}). Additionally, we report the \textit{latency per query} ({\tt Lat}) to estimate efficiency. The exact costs of model execution, including factors like latency, FLOPs, and energy consumption, may vary and are influenced by factors such as prompt templates, hardware capabilities, and network connectivity, especially in LLM inference scenarios. To ensure a fair comparison, we record the inference latency of each LLM under uniform conditions using single A100 GPUs. The individual latencies for each LLM are detailed in Appendix Table \ref{tab:llm_lat}.


\section{Results and Discussion}
\label{sec:results}

\begin{table*}[!t]
\centering
\resizebox{0.8\textwidth}{!}{%
\begin{tabular}{@{}ll|cc|cc@{}}
\hline \hline
\multicolumn{2}{l|}{\multirow{2}{*}{\textbf{Models / Setups}}} & \multicolumn{2}{c|}{\textbf{GSM8K}} & \multicolumn{2}{c}{\textbf{MMLU}} \\ \cline{3-6} 
\multicolumn{2}{l|}{} & \textbf{{\tt Acc} ($\uparrow$)} & \textbf{{\tt Lat} ($\downarrow$)} & \textbf{{\tt Acc} ($\uparrow$)} & \textbf{{\tt Lat} ($\downarrow$)} \\ \hline
\multicolumn{2}{l|}{Oracle} & 90.52 & 3.24 & 90.46 & 1.75 \\ \hline
\multirow{4}{*}{Baseline} & Random & 69.49 & 9.65 & 58.20 & 8.27 \\ \cline{2-6}
 & LLM-Blender \cite{jiang-etal-2023-llm} & 75.28 & 19.00 & 60.27 & 16.40 \\ \cline{2-6} 
 & All LLMs \citep{li2024more} & 76.04 & 19.00 & 60.92 & 16.40 \\ \cline{2-6} 
 & Top-\textit{s} LLMs  & 77.48 & 19.00 & 65.75 & 16.40 \\ \cline{2-6}
 \cline{1-6}
\multirow{3}{*}{\textsc{Select}LLM} & {\tt MLC} +  \textsc{LabelledMaxConf} & 75.66 & 14.69 & 65.68  & 4.78 \\ \cline{2-6} 
 & {\tt MLC} + \textsc{MaxConf} & 77.48 & 16.50 & 65.68 & 4.78 \\ \cline{2-6} 
 & {\tt MLC} + \textsc{WeightedMaxConf} & \textbf{77.94} & 16.50 & \textbf{65.81} & 4.78 \\ \hline
\textsc{Select}LLM & {\tt MLC} + \textsc{WeightedMaxConf} ($s=1$) & 67.24 & 4.70 & 63.52 & 2.97 \\ \hline
\end{tabular}%
}
\vspace{-0.2cm}
\caption{Performance of different models on the \texttt{GSM8K} \citep{cobbe2021training} and \texttt{MMLU} \citep{hendryckstest2021} \textit{test} sets. By default, \(s = 4\) is used for \texttt{GSM8K} and \(s = 2\) for \texttt{MMLU}. The value of \(s\) is selected based on experiments on the validation set (see Appendix Section \ref{subsec:lat_vs_acc}). {\tt Acc}: with {\sc maj}@(M $\times$ $L_s$) scores reported in percentage (\%); {\tt Lat}: runtime of $M$ generations for a single query (in seconds); {\tt MLC}: multi-label classifier; $s$: the number of LLMs considered.}
\vspace{-0.5cm}
\label{tab:main_results}
\end{table*}

Table \ref{tab:main_results} presents the performance results for the Oracle, baselines, and the proposed \textsc{SelectLLM} models across both {\tt GSM8K} and {\tt MMLU} datasets. We have also reported respective inference latencies to analyze the efficiency of different models. We make the following major observations:

\paragraph{Performance of baselines:}
We evaluated four baseline models -- \textit{Random}, \textit{All LLMs}, \textit{LLM-Blender}, and \textit{Top-s LLMs} -- to assess the impact of ensembling LLMs. The performance of the \textit{Random} baseline model surpasses that of several individual LLMs reported in Fig.~\ref{tab:llm_performance}, demonstrating the potential of utilizing multiple LLMs. \citet{li2024more} reports that increasing the number of LLMs in the model pool generally increases ensemble performance. Interestingly, however, selecting only the top few LLMs in our experiments performs better than including all LLMs (in {\em all LLMs}). \textit{Thus, the optimal number of LLMs may depend on the diversity of the model pool (i.e., individual capabilities and overall performance on benchmarks)}. The \textit{Top-s LLMs} model proves to be the most effective baseline, outperforming even robust approaches such as LLM-Blender, indicating the benefit of prioritizing the best-performing LLMs. \textit{However, the latency of most baseline models is high (as they must utilize the entire model pool at inference), limiting their real-world practicality.}

\paragraph{The effect of different policies with the}{\sc \textbf{SelectLLM }} \textbf{algorithm:} 
It can be observed that the \textsc{LabelledMaxConf} policy yields the lowest performance on the {\tt GSM8K} dataset. This suboptimal performance may be attributed to the policy's dependence on both the MLC predictions and confidence scores, wherein the classifier predominantly assigns the label {\tt metamath-7b-lm}. Given that {\tt metamath-7b-lm} is a specialized model for mathematics and 88\% of the \textsc{SLData} training data is associated with this label, this reliance skews the predictions. However, this effect is minimal for the {\tt MMLU} dataset, where the label distribution across LLMs is more balanced. This limitation is addressed by the \textsc{MaxConf} and \textsc{WeightedMaxConf}, which relax the constraint on MLC label prediction and only operate on confidence scores. This allows models to incorporate other LLMs and push the performance, particularly for the {\tt GSM8K} data. Mathematical tuning in \textsc{WeightedMaxConf} allows policies to select LLMs more effectively and improve the scores. \textit{Overall, the \textsc{WeightedMaxConf} policy emerged as the best performing, with a slight edge over its closest competitor, the \textsc{MaxConf} policy.}

\paragraph{\sc \textbf{SelectLLM}}\textbf{ vs. baseline models:} The proposed \textsc{SelectLLM} model, utilizing the \textsc{WeightedMaxConf} policy, demonstrates superior performance compared to the \textit{Random}, \textit{All LLMs}, and \textit{LLM-Blender} baselines. Additionally, \textsc{SelectLLM} maintains competitive accuracy relative to the \textit{Top-$s$ LLMs} baseline while achieving significantly lower latency. Specifically, it reduces latency by 13\% for the {\tt GSM8K} dataset and by 70\% for the {\tt MMLU} dataset compared to all baselines (except \textit{Random}). \textit{This substantial efficiency gain underscores the effectiveness of \textsc{SelectLLM} in balancing performance and operational efficiency. }

\paragraph{Query awareness analysis of \textsc{SelectLLM}:} With this analysis, we aim to understand \textit{how the distribution of selected LLMs changes as more LLMs are selected (with increasing values of $s$)} in \textsc{SelectLLM} algorithm. Figure \ref{fig:subsetllm_dist} in the Appendix presents such distribution. For both datasets, in the top-1 and top-2 subsets, most of the queries are directed to the best-performing LLMs. However, as the subset size increases, the dominance of the top-performing models diminishes, leading to a more uniform distribution where queries are routed towards more LLMs to boost the performance. This indicates the input query awareness of the \textsc{SelectLLM} model, which is adept at assigning a suitable set of LLMs for the input query.

\paragraph{Prediction distribution analysis:} 
Appendix Figure \ref{fig:ques_dist} presents the distribution of the number of input queries with correct answer predictions using the top-3 individual LLMs and models with the \textsc{SelectLLM} algorithm across both datasets. It can be observed (count from the right column) that the proposed model is able to correctly provide answers to input queries compared to other individual LLMs, which supports the performance gain reported in Table \ref{tab:main_results}. The distribution also indicates (count from the bottom) that the proposed model utilizes the capabilities of multiple individual LLMs to extract the correct answers. Additionally, Figure \ref{fig:mmlu_subject_analysis} illustrates the subject-wise performance differences between \textsc{SelectLLM} and the best performing individual LLM (i.e., {\tt gemma-7b-lm}) for {\tt MMLU}. It can be observed that the proposed model shows substantial gains in the majority of subjects, while performing slightly worse for a few of them, showing scope for improvement.

\paragraph{Upper-bound and performance gap with \textsc{SelectLLM}:} We measure the maximum performance achievable by \textsc{SelectLLM} with the \textsc{WeightedMaxConf} policy by establishing an upper-bound performance. Specifically, we compute potential subsets based on the labels predicted by the MLC for each query $q_i$. A query is considered solved if at least one subset yields a correct answer. 
The upper bound scores for GSM8K and MMLU are 78.77 and 76.20, respectively. This reveals the following insights: (1) the classifier's performance is constrained, with weighted F1 scores of 0.71 for {\tt GSM8K} and 0.68 for {\tt MMLU}, due to limited training data (approximately 7K for {\tt GSM8K} and 14K for {\tt MMLU}), likely contributing to the performance gap between \textsc{SelectLLM}  and Oracle. \textit{Therefore, augmenting the training data or improving the classifier model could enhance scores.} (2) A performance disparity between the best policy and the upper bound suggests the potential for developing better policies (specifically for {\tt MMLU}). \textit{However, since policies rely on the classifier's confidence scores, enhancing the classifier could also bridge this gap.} We will investigate this performance gap from a linguistic perspective in Section \ref{sec:intpgap}.

\paragraph{Ablation studies:} \textit{Is the proposed algorithm effective with different LLMs pool sizes (i.e., value of $k$)?} To investigate this question, we conduct ablation studies considering various LLM pool sizes, i.e., $k = {1,...,6}$, for the {\tt GSM8K} dataset.  We examine two extreme settings: pools with top-$k$ and bottom-$k$ LLMs based on individual LLM performance. This encompasses many configurations as LLMs with similar or different performances may be present in the LLM pool. We also compare this with closet strong \textit{Top-$s$ LLM} baseline. The results are shown in the Appendix Figure \ref{fig:ablation_study}. We observe that even with different $k$ values across both top-$k$ and bottom-$k$ setups, the proposed \textsc{SelectLLM} outperforms (in terms of accuracy) the \texttt{Top-$s$ LLM} baseline. \textit{This indicates that the proposed approach is LLM pool-agnostic}. Moreover, as the number of $k$ values increases, latency becomes a factor: for larger pool sizes, the latency for \textsc{SelectLLM} is much lower than for \texttt{Top-$s$ LLM}. Similar results are observed with the {\tt MMLU} dataset.

\begin{figure*}[!htb]
    \centering
    \includegraphics[width=\linewidth]{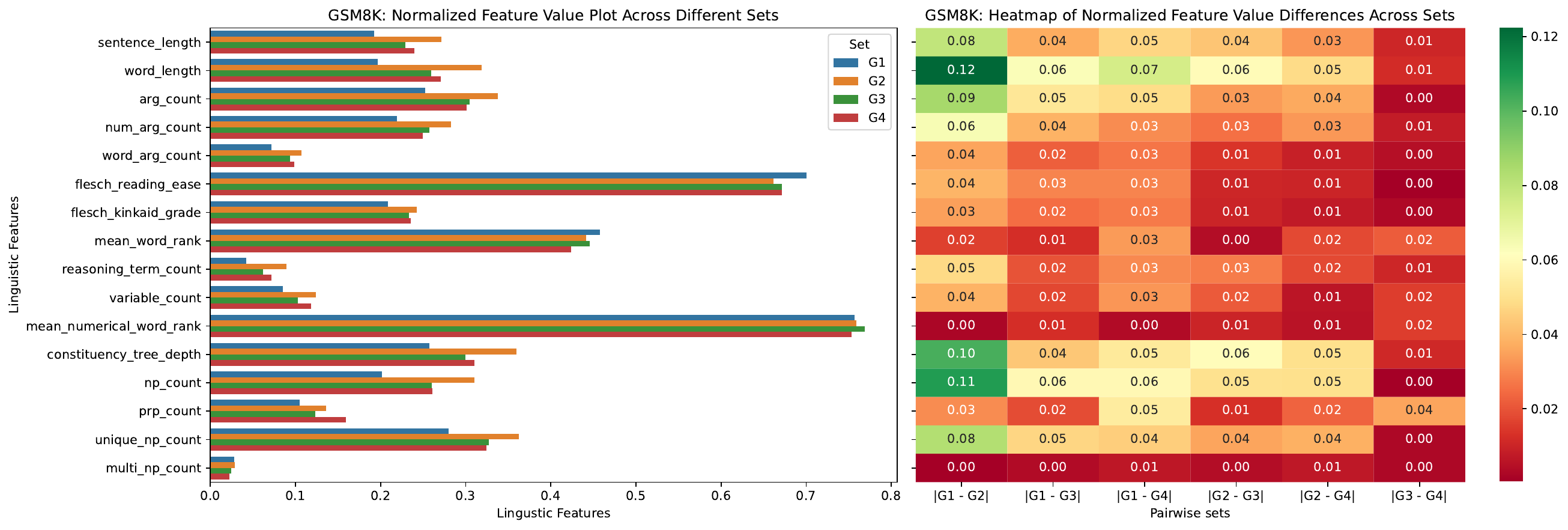}
    \vspace{-0.2cm}
    \caption{Quantitative Analysis ({\tt GSM8K}): Distribution of normalized linguistic feature values (left) and the pairwise absolute difference in normalized feature values between different sets (right).}
        \vspace{-0.5cm}
    \label{fig:quantanalysis}
\end{figure*}

\begin{figure*}[]
    \centering
    \begin{subfigure}[t]{0.5\textwidth}
        \centering
        \includegraphics[height=1.5in]{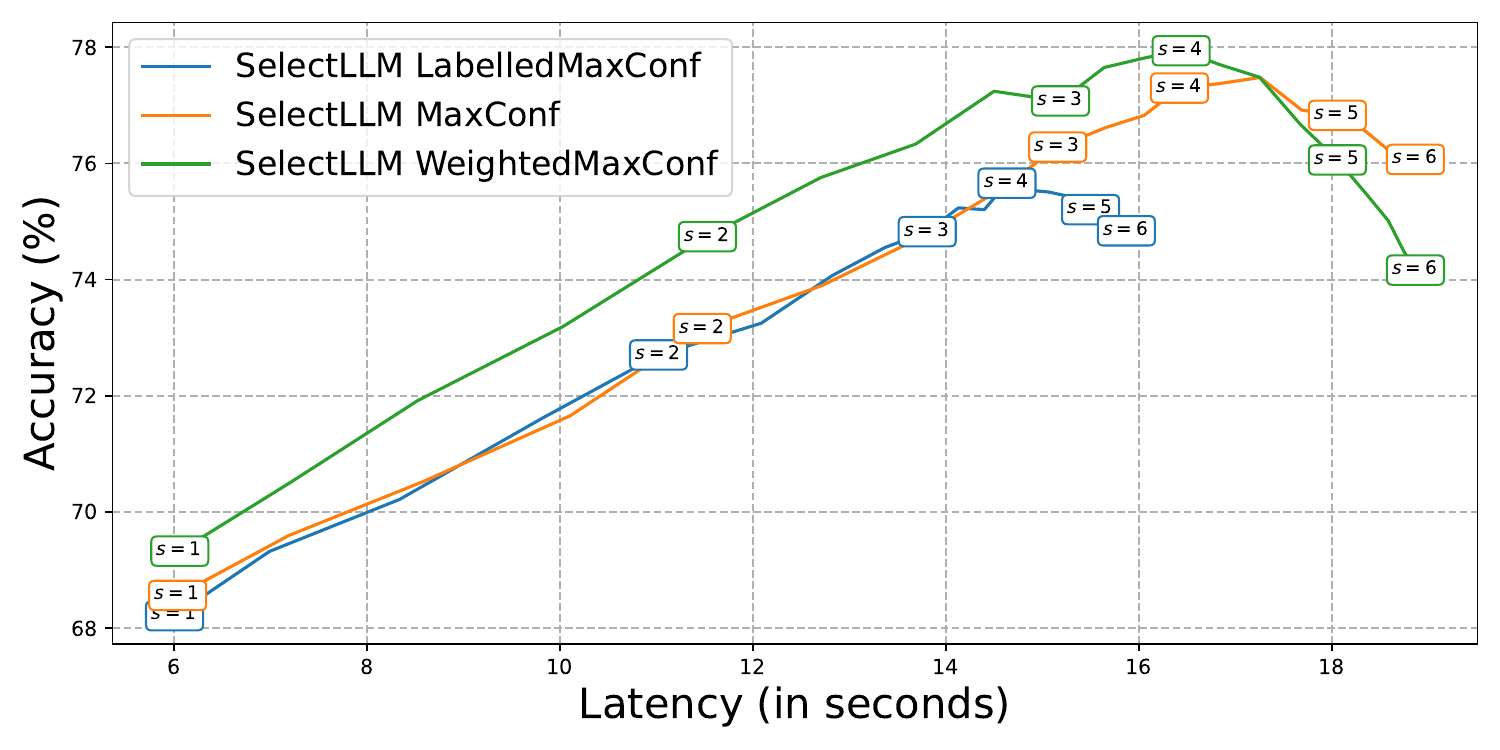}
    \end{subfigure}%
    ~ 
    \begin{subfigure}[t]{0.5\textwidth}
        \centering
        \includegraphics[height=1.5in]{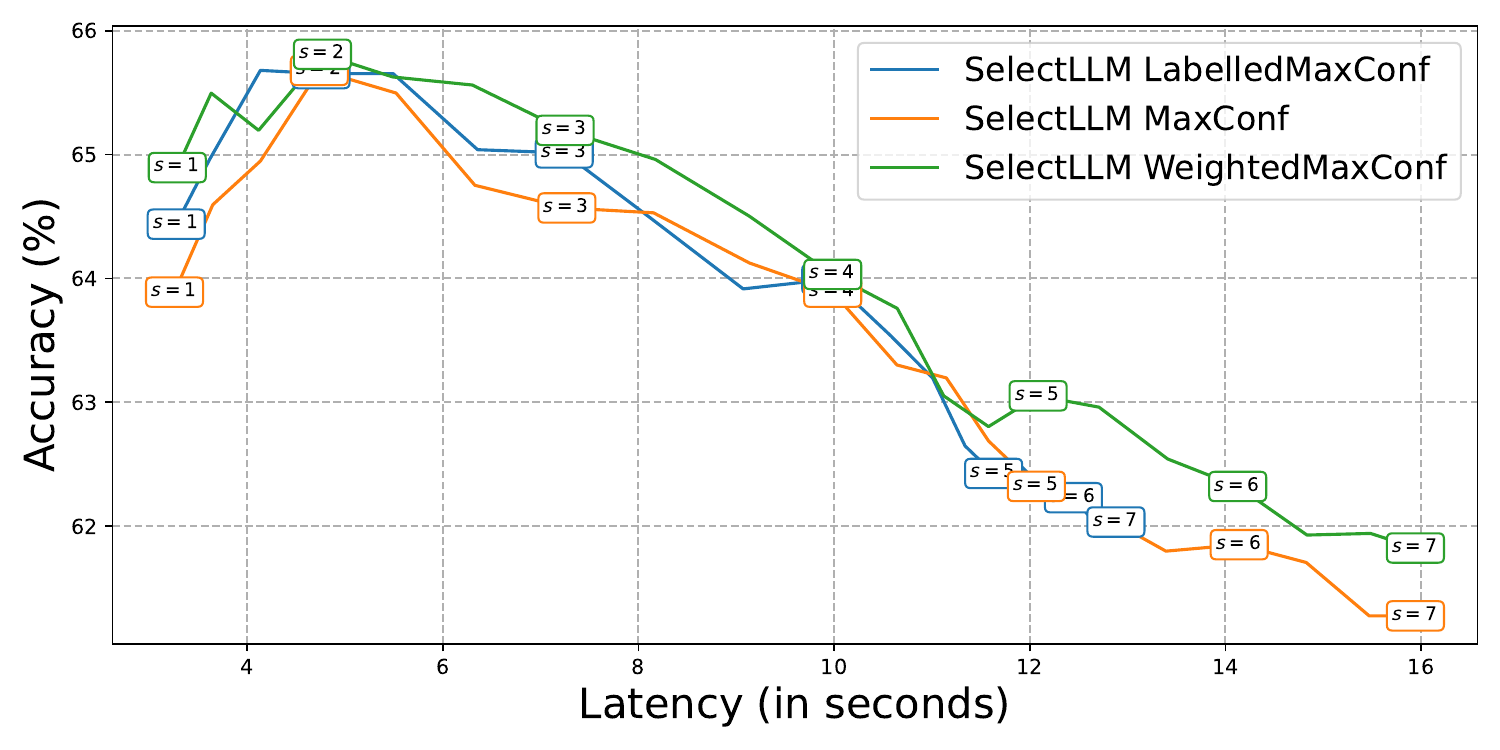}
    \end{subfigure}
    \vspace{-0.7cm}
    \caption{Latency vs. accuracy vs. $s$-values plots for different {\sc Select}LLM policies on {\tt GSM8K} (left) and {\tt MMLU} (right) development sets. $s$-value: total number of LLMs selected with \textsc{SelectLLM} model.}
    \vspace{-0.3cm}
\label{fig:lat_vs_acc}
\end{figure*}

\paragraph{Latency vs. Accuracy vs. $s$-Value:}
\label{subsec:lat_vs_acc}
Figure \ref{fig:lat_vs_acc} shows the relationship between latency, accuracy, and $s$-values for different \textsc{SelectLLM} policies across both datasets along with the corresponding development sets. \textsc{WeightedMaxConf} consistently outperforms both \textsc{MaxConf} and \textsc{Random} in terms of accuracy and latency for most $s$-values (except for 5 and 6 on {\tt GSM8K}). This highlights the superiority of \textsc{WeightedMaxConf}, maintaining its effectiveness even with a small number of LLMs. A larger number of LLMs and higher latency are required for {\tt GSM8K}, whereas a lower $s$-value and latency suffice for {\tt MMLU} to achieve high accuracy. \textit{Selecting 4 LLMs for {\tt GSM8K} and 2 for {\tt MMLU} yields optimal performance, demonstrating the effectiveness and efficiency of \textsc{SelectLLM} with \textsc{WeightedMaxConf}.} Using this optimal value of $s$ results on the test set are reported in Table \ref{tab:main_results}.

Additionally, we also report the performance of \textsc{SelectLLM} with \textsc{WeightedMaxConf} for $s=1$ in Table \ref{tab:main_results} to understand the impact of selecting the best query-aware LLM. The results indicate that there is a large performance gap when selecting multiple LLMs, underscoring that ensembling is a promising direction.

\begin{table*}[]
\centering
\resizebox{\textwidth}{!}{%
\begin{tabular}{ccccccccccccc}
\toprule[2pt]
\textbf{Set} & \textbf{LLM} & \textbf{frequency terms} & \textbf{time duration units} & \textbf{age units} & \textbf{small numbers} & \textbf{quantifiers} & \textbf{ordinals} & \textbf{fractional values} & \textbf{rates and ratios} & \textbf{named entities} & \textbf{other units} & \textbf{groups} \\ \midrule[1pt]
\multirow{6}{*}{\textbf{(G1) Solved by All}} & gemma-7b-it & \cellcolor[HTML]{B30D26}{\color[HTML]{FFFFFF} \texttt{-0.938}} & \cellcolor[HTML]{CA2427}{\color[HTML]{FFFFFF} \texttt{-0.845}} & \cellcolor[HTML]{FEE593}{\color[HTML]{000000} \texttt{-0.167}} & \cellcolor[HTML]{A50026}{\color[HTML]{FFFFFF} \texttt{-1.0}} & \cellcolor[HTML]{A90426}{\color[HTML]{FFFFFF} \texttt{-0.982}} & \cellcolor[HTML]{FFF2AA}{\color[HTML]{000000} \texttt{-0.084}} & \cellcolor[HTML]{FEEA9B}{\color[HTML]{000000} \texttt{-0.138}} & \cellcolor[HTML]{FDFEBC}{\color[HTML]{000000} \texttt{0.014}} & \cellcolor[HTML]{006837}{\color[HTML]{FFFFFF} \texttt{1.0}} & \cellcolor[HTML]{F1F9AC}{\color[HTML]{000000} \texttt{0.073}} & \cellcolor[HTML]{C1E57B}{\color[HTML]{000000} \texttt{0.296}} \\ \cline{2-2}
 & gemma-7b-lm& \cellcolor[HTML]{B50F26}{\color[HTML]{FFFFFF} \texttt{-0.936}} & \cellcolor[HTML]{57B65F}{\color[HTML]{000000} \texttt{0.634}} & \cellcolor[HTML]{D5ED88}{\color[HTML]{000000} \texttt{0.214}} & \cellcolor[HTML]{006837}{\color[HTML]{FFFFFF} \texttt{1.0}} & \cellcolor[HTML]{006837}{\color[HTML]{FFFFFF} \texttt{1.0}} & \cellcolor[HTML]{E2F397}{\color[HTML]{000000} \texttt{0.152}} & \cellcolor[HTML]{2DA155}{\color[HTML]{FFFFFF} \texttt{0.743}} & \cellcolor[HTML]{DDF191}{\color[HTML]{000000} \texttt{0.172}} & \cellcolor[HTML]{006837}{\color[HTML]{FFFFFF} \texttt{1.0}} & \cellcolor[HTML]{3FAA59}{\color[HTML]{000000} \texttt{0.699}} & \cellcolor[HTML]{4BB05C}{\color[HTML]{000000} \texttt{0.669}} \\ \cline{2-2}
 & llama2-13b-chat & \cellcolor[HTML]{A70226}{\color[HTML]{FFFFFF} \texttt{-0.988}} & \cellcolor[HTML]{E44C34}{\color[HTML]{FFFFFF} \texttt{-0.706}} & \cellcolor[HTML]{FFF0A6}{\color[HTML]{000000} \texttt{-0.095}} & \cellcolor[HTML]{A50026}{\color[HTML]{FFFFFF} \texttt{-0.999}} & \cellcolor[HTML]{CC2627}{\color[HTML]{FFFFFF} \texttt{-0.837}} & \cellcolor[HTML]{FFFAB6}{\color[HTML]{000000} \texttt{-0.038}} & \cellcolor[HTML]{FED07E}{\color[HTML]{000000} \texttt{-0.264}} & \cellcolor[HTML]{FEEFA3}{\color[HTML]{000000} \texttt{-0.107}} & \cellcolor[HTML]{006837}{\color[HTML]{FFFFFF} \texttt{1.0}} & \cellcolor[HTML]{E5F49B}{\color[HTML]{000000} \texttt{0.135}} & \cellcolor[HTML]{73C264}{\color[HTML]{000000} \texttt{0.555}} \\ \cline{2-2}
 & metamath-7b-lm& \cellcolor[HTML]{F2FAAE}{\color[HTML]{000000} \texttt{0.065}} & \cellcolor[HTML]{04703B}{\color[HTML]{FFFFFF} \texttt{0.962}} & \cellcolor[HTML]{B7E075}{\color[HTML]{000000} \texttt{0.33}} & \cellcolor[HTML]{006837}{\color[HTML]{FFFFFF} \texttt{1.0}} & \cellcolor[HTML]{006837}{\color[HTML]{FFFFFF} \texttt{1.0}} & \cellcolor[HTML]{C5E67E}{\color[HTML]{000000} \texttt{0.279}} & \cellcolor[HTML]{0B7D42}{\color[HTML]{FFFFFF} \texttt{0.909}} & \cellcolor[HTML]{ADDC6F}{\color[HTML]{000000} \texttt{0.372}} & \cellcolor[HTML]{006837}{\color[HTML]{FFFFFF} \texttt{1.0}} & \cellcolor[HTML]{2AA054}{\color[HTML]{FFFFFF} \texttt{0.753}} & \cellcolor[HTML]{63BC62}{\color[HTML]{000000} \texttt{0.607}} \\ \cline{2-2}
 & mistral-7b-it & \cellcolor[HTML]{A50026}{\color[HTML]{FFFFFF} \texttt{-0.993}} & \cellcolor[HTML]{F7844E}{\color[HTML]{000000} \texttt{-0.531}} & \cellcolor[HTML]{FFF6B0}{\color[HTML]{000000} \texttt{-0.058}} & \cellcolor[HTML]{CC2627}{\color[HTML]{FFFFFF} \texttt{-0.842}} & \cellcolor[HTML]{026C39}{\color[HTML]{FFFFFF} \texttt{0.98}} & \cellcolor[HTML]{FFFCBA}{\color[HTML]{000000} \texttt{-0.023}} & \cellcolor[HTML]{DAF08D}{\color[HTML]{000000} \texttt{0.195}} & \cellcolor[HTML]{F5FBB2}{\color[HTML]{000000} \texttt{0.049}} & \cellcolor[HTML]{006837}{\color[HTML]{FFFFFF} \texttt{1.0}} & \cellcolor[HTML]{AFDD70}{\color[HTML]{000000} \texttt{0.364}} & \cellcolor[HTML]{7FC866}{\color[HTML]{000000} \texttt{0.519}} \\ \cline{2-2}
 & mistral-7b-lm & \cellcolor[HTML]{A50026}{\color[HTML]{FFFFFF} \texttt{-0.994}} & \cellcolor[HTML]{FB9D59}{\color[HTML]{000000} \texttt{-0.452}} & \cellcolor[HTML]{FFFDBC}{\color[HTML]{000000} \texttt{-0.009}} & \cellcolor[HTML]{006837}{\color[HTML]{FFFFFF} \texttt{1.0}} & \cellcolor[HTML]{128A49}{\color[HTML]{FFFFFF} \texttt{0.857}} & \cellcolor[HTML]{FFFDBC}{\color[HTML]{000000} \texttt{-0.009}} & \cellcolor[HTML]{EFF8AA}{\color[HTML]{000000} \texttt{0.085}} & \cellcolor[HTML]{FFFEBE}{\color[HTML]{000000} \texttt{-0.002}} & \cellcolor[HTML]{006837}{\color[HTML]{FFFFFF} \texttt{1.0}} & \cellcolor[HTML]{B9E176}{\color[HTML]{000000} \texttt{0.327}} & \cellcolor[HTML]{8ECF67}{\color[HTML]{000000} \texttt{0.472}} \\ \midrule[1pt]
\multirow{6}{*}{\textbf{(G2) Solved by SelectLLM}} & gemma-7b-it & \cellcolor[HTML]{A50026}{\color[HTML]{FFFFFF} \texttt{-1.0}} & \cellcolor[HTML]{A50026}{\color[HTML]{FFFFFF} \texttt{-0.998}} & \cellcolor[HTML]{B10B26}{\color[HTML]{FFFFFF} \texttt{-0.949}} & \cellcolor[HTML]{A50026}{\color[HTML]{FFFFFF} \texttt{-1.0}} & \cellcolor[HTML]{A50026}{\color[HTML]{FFFFFF} \texttt{-1.0}} & \cellcolor[HTML]{FDC574}{\color[HTML]{000000} \texttt{-0.312}} & \cellcolor[HTML]{A50026}{\color[HTML]{FFFFFF} \texttt{-0.996}} & \cellcolor[HTML]{F36B42}{\color[HTML]{000000} \texttt{-0.607}} & \cellcolor[HTML]{006837}{\color[HTML]{FFFFFF} \texttt{1.0}} & \cellcolor[HTML]{FEDA86}{\color[HTML]{000000} \texttt{-0.225}} & \cellcolor[HTML]{0B7D42}{\color[HTML]{FFFFFF} \texttt{0.911}} \\ \cline{2-2}
 & gemma-7b-lm& \cellcolor[HTML]{A50026}{\color[HTML]{FFFFFF} \texttt{-1.0}} & \cellcolor[HTML]{219C52}{\color[HTML]{FFFFFF} \texttt{0.775}} & \cellcolor[HTML]{006837}{\color[HTML]{FFFFFF} \texttt{0.999}} & \cellcolor[HTML]{006837}{\color[HTML]{FFFFFF} \texttt{1.0}} & \cellcolor[HTML]{006837}{\color[HTML]{FFFFFF} \texttt{1.0}} & \cellcolor[HTML]{128A49}{\color[HTML]{FFFFFF} \texttt{0.854}} & \cellcolor[HTML]{FECE7C}{\color[HTML]{000000} \texttt{-0.268}} & \cellcolor[HTML]{8ECF67}{\color[HTML]{000000} \texttt{0.472}} & \cellcolor[HTML]{006837}{\color[HTML]{FFFFFF} \texttt{1.0}} & \cellcolor[HTML]{0B7D42}{\color[HTML]{FFFFFF} \texttt{0.913}} & \cellcolor[HTML]{016A38}{\color[HTML]{FFFFFF} \texttt{0.988}} \\ \cline{2-2}
 & llama2-13b-chat & \cellcolor[HTML]{A50026}{\color[HTML]{FFFFFF} \texttt{-1.0}} & \cellcolor[HTML]{A50026}{\color[HTML]{FFFFFF} \texttt{-0.994}} & \cellcolor[HTML]{FFF5AE}{\color[HTML]{000000} \texttt{-0.068}} & \cellcolor[HTML]{A50026}{\color[HTML]{FFFFFF} \texttt{-1.0}} & \cellcolor[HTML]{A50026}{\color[HTML]{FFFFFF} \texttt{-1.0}} & \cellcolor[HTML]{FBFDBA}{\color[HTML]{000000} \texttt{0.021}} & \cellcolor[HTML]{A50026}{\color[HTML]{FFFFFF} \texttt{-0.998}} & \cellcolor[HTML]{E34933}{\color[HTML]{FFFFFF} \texttt{-0.717}} & \cellcolor[HTML]{006837}{\color[HTML]{FFFFFF} \texttt{1.0}} & \cellcolor[HTML]{EFF8AA}{\color[HTML]{000000} \texttt{0.081}} & \cellcolor[HTML]{026C39}{\color[HTML]{FFFFFF} \texttt{0.977}} \\ \cline{2-2}
 & metamath-7b-lm& \cellcolor[HTML]{96D268}{\color[HTML]{000000} \texttt{0.45}} & \cellcolor[HTML]{006837}{\color[HTML]{FFFFFF} \texttt{1.0}} & \cellcolor[HTML]{006837}{\color[HTML]{FFFFFF} \texttt{1.0}} & \cellcolor[HTML]{006837}{\color[HTML]{FFFFFF} \texttt{1.0}} & \cellcolor[HTML]{006837}{\color[HTML]{FFFFFF} \texttt{1.0}} & \cellcolor[HTML]{06733D}{\color[HTML]{FFFFFF} \texttt{0.945}} & \cellcolor[HTML]{006837}{\color[HTML]{FFFFFF} \texttt{0.995}} & \cellcolor[HTML]{0D8044}{\color[HTML]{FFFFFF} \texttt{0.897}} & \cellcolor[HTML]{006837}{\color[HTML]{FFFFFF} \texttt{1.0}} & \cellcolor[HTML]{04703B}{\color[HTML]{FFFFFF} \texttt{0.964}} & \cellcolor[HTML]{04703B}{\color[HTML]{FFFFFF} \texttt{0.968}} \\ \cline{2-2}
 & mistral-7b-it & \cellcolor[HTML]{A50026}{\color[HTML]{FFFFFF} \texttt{-1.0}} & \cellcolor[HTML]{A70226}{\color[HTML]{FFFFFF} \texttt{-0.988}} & \cellcolor[HTML]{F2FAAE}{\color[HTML]{000000} \texttt{0.07}} & \cellcolor[HTML]{A50026}{\color[HTML]{FFFFFF} \texttt{-1.0}} & \cellcolor[HTML]{006837}{\color[HTML]{FFFFFF} \texttt{1.0}} & \cellcolor[HTML]{B5DF74}{\color[HTML]{000000} \texttt{0.34}} & \cellcolor[HTML]{A50026}{\color[HTML]{FFFFFF} \texttt{-0.996}} & \cellcolor[HTML]{FDC372}{\color[HTML]{000000} \texttt{-0.314}} & \cellcolor[HTML]{006837}{\color[HTML]{FFFFFF} \texttt{1.0}} & \cellcolor[HTML]{84CA66}{\color[HTML]{000000} \texttt{0.504}} & \cellcolor[HTML]{026C39}{\color[HTML]{FFFFFF} \texttt{0.983}} \\ \cline{2-2}
 & mistral-7b-lm & \cellcolor[HTML]{A50026}{\color[HTML]{FFFFFF} \texttt{-1.0}} & \cellcolor[HTML]{A70226}{\color[HTML]{FFFFFF} \texttt{-0.984}} & \cellcolor[HTML]{96D268}{\color[HTML]{000000} \texttt{0.448}} & \cellcolor[HTML]{A50026}{\color[HTML]{FFFFFF} \texttt{-1.0}} & \cellcolor[HTML]{016A38}{\color[HTML]{FFFFFF} \texttt{0.989}} & \cellcolor[HTML]{B9E176}{\color[HTML]{000000} \texttt{0.322}} & \cellcolor[HTML]{A50026}{\color[HTML]{FFFFFF} \texttt{-0.994}} & \cellcolor[HTML]{F88950}{\color[HTML]{000000} \texttt{-0.513}} & \cellcolor[HTML]{006837}{\color[HTML]{FFFFFF} \texttt{1.0}} & \cellcolor[HTML]{8ECF67}{\color[HTML]{000000} \texttt{0.472}} & \cellcolor[HTML]{04703B}{\color[HTML]{FFFFFF} \texttt{0.966}} \\ \midrule[1pt]
\multirow{6}{*}{\textbf{(G3) Solved by Oracle}} & gemma-7b-it & \cellcolor[HTML]{E95538}{\color[HTML]{FFFFFF} \texttt{-0.674}} & \cellcolor[HTML]{DB382B}{\color[HTML]{FFFFFF} \texttt{-0.77}} & \cellcolor[HTML]{FDBF6F}{\color[HTML]{000000} \texttt{-0.334}} & \cellcolor[HTML]{A50026}{\color[HTML]{FFFFFF} \texttt{-1.0}} & \cellcolor[HTML]{A90426}{\color[HTML]{FFFFFF} \texttt{-0.977}} & \cellcolor[HTML]{FECE7C}{\color[HTML]{000000} \texttt{-0.272}} & \cellcolor[HTML]{FEE491}{\color[HTML]{000000} \texttt{-0.176}} & \cellcolor[HTML]{F5FBB2}{\color[HTML]{000000} \texttt{0.05}} & \cellcolor[HTML]{A50026}{\color[HTML]{FFFFFF} \texttt{-1.0}} & \cellcolor[HTML]{FECC7B}{\color[HTML]{000000} \texttt{-0.281}} & \cellcolor[HTML]{F8FCB6}{\color[HTML]{000000} \texttt{0.037}} \\ \cline{2-2}
 & gemma-7b-lm& \cellcolor[HTML]{E44C34}{\color[HTML]{FFFFFF} \texttt{-0.707}} & \cellcolor[HTML]{DCF08F}{\color[HTML]{000000} \texttt{0.185}} & \cellcolor[HTML]{CDEA83}{\color[HTML]{000000} \texttt{0.248}} & \cellcolor[HTML]{006837}{\color[HTML]{FFFFFF} \texttt{0.999}} & \cellcolor[HTML]{0C7F43}{\color[HTML]{FFFFFF} \texttt{0.902}} & \cellcolor[HTML]{B1DE71}{\color[HTML]{000000} \texttt{0.357}} & \cellcolor[HTML]{B3DF72}{\color[HTML]{000000} \texttt{0.351}} & \cellcolor[HTML]{B7E075}{\color[HTML]{000000} \texttt{0.332}} & \cellcolor[HTML]{006837}{\color[HTML]{FFFFFF} \texttt{1.0}} & \cellcolor[HTML]{DCF08F}{\color[HTML]{000000} \texttt{0.183}} & \cellcolor[HTML]{E6F59D}{\color[HTML]{000000} \texttt{0.129}} \\ \cline{2-2}
 & llama2-13b-chat & \cellcolor[HTML]{CE2827}{\color[HTML]{FFFFFF} \texttt{-0.836}} & \cellcolor[HTML]{E65036}{\color[HTML]{FFFFFF} \texttt{-0.689}} & \cellcolor[HTML]{FEE28F}{\color[HTML]{000000} \texttt{-0.183}} & \cellcolor[HTML]{A50026}{\color[HTML]{FFFFFF} \texttt{-1.0}} & \cellcolor[HTML]{B30D26}{\color[HTML]{FFFFFF} \texttt{-0.944}} & \cellcolor[HTML]{FEEDA1}{\color[HTML]{000000} \texttt{-0.116}} & \cellcolor[HTML]{FFF8B4}{\color[HTML]{000000} \texttt{-0.043}} & \cellcolor[HTML]{F7FCB4}{\color[HTML]{000000} \texttt{0.042}} & \cellcolor[HTML]{016A38}{\color[HTML]{FFFFFF} \texttt{0.988}} & \cellcolor[HTML]{FED683}{\color[HTML]{000000} \texttt{-0.241}} & \cellcolor[HTML]{E8F59F}{\color[HTML]{000000} \texttt{0.12}} \\ \cline{2-2}
 & metamath-7b-lm& \cellcolor[HTML]{ECF7A6}{\color[HTML]{000000} \texttt{0.102}} & \cellcolor[HTML]{17934E}{\color[HTML]{FFFFFF} \texttt{0.818}} & \cellcolor[HTML]{91D068}{\color[HTML]{000000} \texttt{0.466}} & \cellcolor[HTML]{006837}{\color[HTML]{FFFFFF} \texttt{1.0}} & \cellcolor[HTML]{006837}{\color[HTML]{FFFFFF} \texttt{0.999}} & \cellcolor[HTML]{66BD63}{\color[HTML]{000000} \texttt{0.599}} & \cellcolor[HTML]{5DB961}{\color[HTML]{000000} \texttt{0.62}} & \cellcolor[HTML]{AFDD70}{\color[HTML]{000000} \texttt{0.367}} & \cellcolor[HTML]{006837}{\color[HTML]{FFFFFF} \texttt{1.0}} & \cellcolor[HTML]{8ECF67}{\color[HTML]{000000} \texttt{0.475}} & \cellcolor[HTML]{D9EF8B}{\color[HTML]{000000} \texttt{0.196}} \\ \cline{2-2}
 & mistral-7b-it & \cellcolor[HTML]{C62027}{\color[HTML]{FFFFFF} \texttt{-0.864}} & \cellcolor[HTML]{EE613E}{\color[HTML]{000000} \texttt{-0.635}} & \cellcolor[HTML]{FEE18D}{\color[HTML]{000000} \texttt{-0.194}} & \cellcolor[HTML]{A50026}{\color[HTML]{FFFFFF} \texttt{-1.0}} & \cellcolor[HTML]{F67F4B}{\color[HTML]{000000} \texttt{-0.542}} & \cellcolor[HTML]{FFFBB8}{\color[HTML]{000000} \texttt{-0.027}} & \cellcolor[HTML]{FFFDBC}{\color[HTML]{000000} \texttt{-0.01}} & \cellcolor[HTML]{D9EF8B}{\color[HTML]{000000} \texttt{0.195}} & \cellcolor[HTML]{006837}{\color[HTML]{FFFFFF} \texttt{0.998}} & \cellcolor[HTML]{FEE08B}{\color[HTML]{000000} \texttt{-0.197}} & \cellcolor[HTML]{EBF7A3}{\color[HTML]{000000} \texttt{0.102}} \\ \cline{2-2}
 & mistral-7b-lm & \cellcolor[HTML]{C21C27}{\color[HTML]{FFFFFF} \texttt{-0.88}} & \cellcolor[HTML]{E75337}{\color[HTML]{FFFFFF} \texttt{-0.681}} & \cellcolor[HTML]{FEE18D}{\color[HTML]{000000} \texttt{-0.189}} & \cellcolor[HTML]{A50026}{\color[HTML]{FFFFFF} \texttt{-1.0}} & \cellcolor[HTML]{E14430}{\color[HTML]{FFFFFF} \texttt{-0.734}} & \cellcolor[HTML]{FFFDBC}{\color[HTML]{000000} \texttt{-0.013}} & \cellcolor[HTML]{FDFEBC}{\color[HTML]{000000} \texttt{0.013}} & \cellcolor[HTML]{E9F6A1}{\color[HTML]{000000} \texttt{0.115}} & \cellcolor[HTML]{006837}{\color[HTML]{FFFFFF} \texttt{1.0}} & \cellcolor[HTML]{FEE593}{\color[HTML]{000000} \texttt{-0.166}} & \cellcolor[HTML]{EEF8A8}{\color[HTML]{000000} \texttt{0.09}} \\ \midrule[1pt]
\multirow{6}{*}{\textbf{(G4) Solved by None}} & gemma-7b-it & \cellcolor[HTML]{ED5F3C}{\color[HTML]{000000} \texttt{-0.646}} & \cellcolor[HTML]{E75337}{\color[HTML]{FFFFFF} \texttt{-0.684}} & \cellcolor[HTML]{D22B27}{\color[HTML]{FFFFFF} \texttt{-0.817}} & \cellcolor[HTML]{A50026}{\color[HTML]{FFFFFF} \texttt{-1.0}} & \cellcolor[HTML]{B10B26}{\color[HTML]{FFFFFF} \texttt{-0.948}} & \cellcolor[HTML]{FECC7B}{\color[HTML]{000000} \texttt{-0.279}} & \cellcolor[HTML]{E34933}{\color[HTML]{FFFFFF} \texttt{-0.713}} & \cellcolor[HTML]{FDC372}{\color[HTML]{000000} \texttt{-0.317}} & \cellcolor[HTML]{A50026}{\color[HTML]{FFFFFF} \texttt{-1.0}} & \cellcolor[HTML]{FEE18D}{\color[HTML]{000000} \texttt{-0.192}} & \cellcolor[HTML]{FFFEBE}{\color[HTML]{000000} \texttt{-0.002}} \\ \cline{2-2}
 & gemma-7b-lm& \cellcolor[HTML]{EF633F}{\color[HTML]{000000} \texttt{-0.629}} & \cellcolor[HTML]{CDEA83}{\color[HTML]{000000} \texttt{0.246}} & \cellcolor[HTML]{E6F59D}{\color[HTML]{000000} \texttt{0.129}} & \cellcolor[HTML]{006837}{\color[HTML]{FFFFFF} \texttt{0.997}} & \cellcolor[HTML]{199750}{\color[HTML]{FFFFFF} \texttt{0.802}} & \cellcolor[HTML]{FFF3AC}{\color[HTML]{000000} \texttt{-0.07}} & \cellcolor[HTML]{FED481}{\color[HTML]{000000} \texttt{-0.245}} & \cellcolor[HTML]{DFF293}{\color[HTML]{000000} \texttt{0.171}} & \cellcolor[HTML]{006837}{\color[HTML]{FFFFFF} \texttt{1.0}} & \cellcolor[HTML]{E9F6A1}{\color[HTML]{000000} \texttt{0.112}} & \cellcolor[HTML]{D9EF8B}{\color[HTML]{000000} \texttt{0.198}} \\ \cline{2-2}
 & llama2-13b-chat & \cellcolor[HTML]{DA362A}{\color[HTML]{FFFFFF} \texttt{-0.776}} & \cellcolor[HTML]{F7844E}{\color[HTML]{000000} \texttt{-0.53}} & \cellcolor[HTML]{DE402E}{\color[HTML]{FFFFFF} \texttt{-0.742}} & \cellcolor[HTML]{A50026}{\color[HTML]{FFFFFF} \texttt{-1.0}} & \cellcolor[HTML]{C41E27}{\color[HTML]{FFFFFF} \texttt{-0.872}} & \cellcolor[HTML]{FED07E}{\color[HTML]{000000} \texttt{-0.259}} & \cellcolor[HTML]{DD3D2D}{\color[HTML]{FFFFFF} \texttt{-0.75}} & \cellcolor[HTML]{FDBF6F}{\color[HTML]{000000} \texttt{-0.332}} & \cellcolor[HTML]{A90426}{\color[HTML]{FFFFFF} \texttt{-0.981}} & \cellcolor[HTML]{FEE593}{\color[HTML]{000000} \texttt{-0.165}} & \cellcolor[HTML]{E9F6A1}{\color[HTML]{000000} \texttt{0.113}} \\ \cline{2-2}
 & metamath-7b-lm& \cellcolor[HTML]{E3F399}{\color[HTML]{000000} \texttt{0.145}} & \cellcolor[HTML]{279F53}{\color[HTML]{FFFFFF} \texttt{0.763}} & \cellcolor[HTML]{128A49}{\color[HTML]{FFFFFF} \texttt{0.857}} & \cellcolor[HTML]{006837}{\color[HTML]{FFFFFF} \texttt{1.0}} & \cellcolor[HTML]{006837}{\color[HTML]{FFFFFF} \texttt{0.992}} & \cellcolor[HTML]{C9E881}{\color[HTML]{000000} \texttt{0.264}} & \cellcolor[HTML]{7DC765}{\color[HTML]{000000} \texttt{0.525}} & \cellcolor[HTML]{82C966}{\color[HTML]{000000} \texttt{0.513}} & \cellcolor[HTML]{006837}{\color[HTML]{FFFFFF} \texttt{1.0}} & \cellcolor[HTML]{BFE47A}{\color[HTML]{000000} \texttt{0.304}} & \cellcolor[HTML]{BFE47A}{\color[HTML]{000000} \texttt{0.304}} \\ \cline{2-2}
 & mistral-7b-it & \cellcolor[HTML]{C82227}{\color[HTML]{FFFFFF} \texttt{-0.858}} & \cellcolor[HTML]{F88950}{\color[HTML]{000000} \texttt{-0.508}} & \cellcolor[HTML]{DE402E}{\color[HTML]{FFFFFF} \texttt{-0.744}} & \cellcolor[HTML]{A50026}{\color[HTML]{FFFFFF} \texttt{-1.0}} & \cellcolor[HTML]{FA9857}{\color[HTML]{000000} \texttt{-0.463}} & \cellcolor[HTML]{FED07E}{\color[HTML]{000000} \texttt{-0.265}} & \cellcolor[HTML]{E0422F}{\color[HTML]{FFFFFF} \texttt{-0.735}} & \cellcolor[HTML]{FEEB9D}{\color[HTML]{000000} \texttt{-0.132}} & \cellcolor[HTML]{C01A27}{\color[HTML]{FFFFFF} \texttt{-0.884}} & \cellcolor[HTML]{FEEB9D}{\color[HTML]{000000} \texttt{-0.127}} & \cellcolor[HTML]{E9F6A1}{\color[HTML]{000000} \texttt{0.111}} \\ \cline{2-2}
 & mistral-7b-lm & \cellcolor[HTML]{CA2427}{\color[HTML]{FFFFFF} \texttt{-0.849}} & \cellcolor[HTML]{FA9656}{\color[HTML]{000000} \texttt{-0.47}} & \cellcolor[HTML]{DD3D2D}{\color[HTML]{FFFFFF} \texttt{-0.753}} & \cellcolor[HTML]{A50026}{\color[HTML]{FFFFFF} \texttt{-1.0}} & \cellcolor[HTML]{F36B42}{\color[HTML]{000000} \texttt{-0.606}} & \cellcolor[HTML]{FEDA86}{\color[HTML]{000000} \texttt{-0.223}} & \cellcolor[HTML]{E34933}{\color[HTML]{FFFFFF} \texttt{-0.712}} & \cellcolor[HTML]{FED683}{\color[HTML]{000000} \texttt{-0.237}} & \cellcolor[HTML]{FFF6B0}{\color[HTML]{000000} \texttt{-0.06}} & \cellcolor[HTML]{FEEC9F}{\color[HTML]{000000} \texttt{-0.117}} & \cellcolor[HTML]{EBF7A3}{\color[HTML]{000000} \texttt{0.103}} \\ \bottomrule[2pt]
\end{tabular}%
}
\caption{Normalized SHAP values for the SelectLLM classifier ({\tt GSM8K}) over questions from the four subsets of the Test split (see Section \ref{sec:intpgap}). Each column represents a related group of Synsets. A negative SHAP value (in red) is detrimental to the predicted solvability of corresponding questions, and a positive value (in green) is beneficial.}
\vspace{-0.2cm}
\label{tab:shap-gsm8k}
\end{table*}

\subsection{Interpretation of Performance Gaps}
\label{sec:intpgap}
We investigate the underlying factors contributing to the performance gap between \textsc{SelectLLM} and the Oracle. To gain insights, we perform both linguistic feature-based quantitative analysis and a SHAP values-based qualitative analysis, examining four subsets (i.e., group) of the test data (numbers are specific to the {\tt GSM8K} dataset): \textbf{G1:} The set of questions correctly solved by all individual LLMs (949 questions), \textbf{G2:} The set of \textit{additional} questions solved only by \textsc{SelectLLM} and the Oracle (46 questions), \textbf{G3:} The set of \textit{additional} questions only solved by the Oracle (168 questions) and \textbf{G4:} The set of questions that could not be solved by any LLM (125 questions).

\paragraph{Quantitative Analysis:} We extracted 16 linguistic features specific to input questions, as proposed by \citet{srivatsa2024naacl} for the {\tt GSM8K} dataset. In Figure \ref{fig:quantanalysis}, we present the feature value distribution (bar plot) and pair-wise value differences (heatmap) across all four subsets. Major insights are: (1) The feature value distributions for G3 and G4 are notably similar, suggesting that questions only solvable by the Oracle exhibit comparable complexity to those not solved by any LLMs. (2) Features such as \textit{length}, \textit{tree depth}, \textit{NP count}, and \textit{argument count} emerge as influential across the different subsets, which aligns with the findings of \citet{rabinovich2023predicting} and \citet{srivatsa2024naacl}. A similar observation holds for the MMLU dataset. Future models should incorporate these features to address the performance gap between \textsc{SelectLLM} and the Oracle.

\paragraph{Qualitative Analysis:} 
We extracted word-level SHAP values\footnote{Some samples of interactive SHAP plots are hosted at \href{https://anonymous.4open.science/r/SelectLLM-3621}{\texttt{https://anonymous.4open.science/r/SelectLLM-3621}}.} from the \textsc{SelectLLM} classifier for {\tt GSM8K}. After mapping key content words to WordNet \cite{miller-1994-wordnet} synsets, we identified synset groups most detrimental to the solvability of each LLM. Table \ref{tab:shap-gsm8k} shows the corresponding SHAP values for {\tt GSM8K}. See Appendix \ref{sec:shap-appendix} for more details. We find that: (1) Not all synset groups affect LLMs uniformly; some are detrimental to one model but beneficial to another (e.g., \textit{small numbers} \& \textit{quantifiers}). \textit{Frequency terms} (e.g., \texttt{``twice"}, \texttt{``thrice"}) consistently reduce solvability for most LLMs, but \texttt{MetaMath} remains resilient. (2) \textit{Gain using \textsc{SelectLLM}}: \textsc{SelectLLM} solves questions with more challenging \textit{fractional values} and \textit{rates and ratios} than individual LLMs. (3) \textit{Gap between \textsc{SelectLLM} and Oracle}: \textit{quantifiers}, \textit{age units}, \textit{other units}, and questions about \textit{groups} like \texttt{"family"} or \texttt{"team"} relatively increase difficulty in G3 compared to G2.  (4) \textit{Solved by None}: G3/G4 rows show that unsolvable questions contain challenging \textit{age units}, \textit{ordinals} (e.g., \texttt{"first"}, \texttt{"third"}), and \textit{named entities}.


\subsection{Out-of-domain (OOD) Generalization}
This section presents a preliminary experiment to validate out-of-domain (OOD) generalization of \textsc{SelectLLM} using the MMLU dataset. As described in Section \ref{sec:llmsample}, the MMLU dataset comprises 57 category-specific subsets based on grade level (high school and college), subject, domain, etc. We conducted two OOD experiments: one at the grade level and another at the subject level. For the grade-level experiment, the model was trained on high school data and evaluated on college-level data. For the subject-level experiment, we created two non-overlapping splits for training and testing. The results, presented in Table \ref{tab:ood_results}, indicate that OOD performance is comparable to in-domain performance, with subject-level generalization yielding even higher accuracy.

\begin{table}[!htb]
    \centering
    \resizebox{0.8\linewidth}{!}{%
    \begin{tabular}{l|c|c}
         \hline 
         \textbf{Setup} & \texttt{\textbf{Acc}} $(\uparrow)$ & \texttt{\textbf{Lat}} $(\downarrow)$ \\ \hline 
         OOD (Grade-level) & 58.97& 4.80 \\
         OOD (Subject-level) & 79.75 & 4.89 \\
         In-domain & 65.81 & 4.78 \\\hline 
    \end{tabular}%
    }
    \caption{\small OOD experiment results with MMLU dataset for \textsc{SelectLLM} with \textsc{WeightedMaxConf} policy and s=2.}
    \label{tab:ood_results}
    \vspace{-0.5cm}
\end{table}

\section{Conclusions and Future Directions}
\label{sec:con}
In this work, we introduce an efficient and novel LLM selection algorithm, \textsc{SelectLLM}, to navigate input queries to the most suitable subset of LLMs from a large pool. \textsc{SelectLLM} employs a multi-label classifier and confidence-based optimal policies to select a lightweight subset of LLMs. The model is evaluated on two challenging reasoning datasets and compared against several strong baseline models. \textsc{SelectLLM} demonstrates superior performance compared to these baseline models, achieving competitive accuracy with a similar subset size of top-performing LLMs, while maintaining significantly lower latency. Despite the promising results, we recognize that further advancements in modeling could bring performance closer to the Oracle benchmark. Our findings provide a robust foundation for future research. A potential avenue for improvement could involve integrating LLM-related and query-related features to enhance the model's query and LLM awareness.



\section*{Limitations} 

We believe that this study provides a useful starting point for optimal subset selection of LLMs from a large pool. However, we acknowledge that there are certain limitations of this work, and addressing these limitations in the future is an important task.

\paragraph{Extension to general generation tasks}
The scope of the current study is limited to two challenging reasoning benchmarks: GSM8K and MMLU. These tasks are framed as generative tasks, where LLMs are required to produce discrete final answers. The study does not focus on general generation tasks such as machine translation, question generation, etc. 

It is well known that the \textit{one model fits all} approach is not universally applicable \cite{huang-etal-2024-one}. For instance, advanced LLMs such as DeepSeek-R1 \cite{deepseekai2025deepseekr1incentivizingreasoningcapability} demonstrate superior performance in reasoning tasks but exhibit comparatively lower effectiveness in general generation tasks \cite{mercer2025brief}. Similar observations hold for OpenAI's and Google's reasoning models. Building on this observation, we posit that SelectLLM serves as a pivotal step toward the efficient selection of an optimal subset of LLMs from a larger pool, balancing both latency and accuracy, with a focus on reasoning tasks. While the current study does not cover general generation tasks, which are left for future work, the primary challenge for such adaptation lies in implementing robust voting mechanisms. These can be modeled in a simpler manner, as proposed in the All LLMs \cite{li2023camel} baseline, to ensure feasibility. Similarly, the idea proposed by \citet{odumakinde2024multilingual} could be adapted for general generation tasks.

\paragraph{Collection of larger training data for multi-label classifier}
Another limitation of the proposed \textsc{SelectLLM} algorithm is the limited availability of training data for the multi-label classifier, with only 7K instances for {\tt GSM8K} and 14K for {\tt MMLU}. This limitation can potentially lead to biased learning. Despite several measures to address this issue, such as weighing labels to counteract label imbalance, conducting extensive optimal hyperparameter searches, experimenting with different sizes of probabilistic and LLM-based models (with RoBERTa performing the best), and obtaining the best checkpoint with the validation set, the performance remains suboptimal. The algorithm achieves a weighted F1 score of 0.71 for {\tt GSM8K} and 0.68 for {\tt MMLU}.

\paragraph{Reduction in invalid answers}
We are able to extract viable answers for 92\% to 95\% of queries across different LLMs, while the remaining queries resulted in invalid/incorrect outputs from the extraction algorithm. These invalid responses can be attributed to two primary factors: (i) limitations in the LLMs' response generation, where outputs are not structured in an extractable format, and (ii)  limitations in the extraction algorithm, which fails to accurately parse the generated text. Similar limitations have also been reported in previous studies \cite{singh2023tree, chen2022program}, particularly for tasks requiring discrete outputs from LLMs. Since the extraction algorithm is the same across all LLMs, the findings are expected to remain the same even if a more effective extraction algorithm is proposed. Future work should prioritize the development of advanced prompting methodologies and optimized LLM architectures that facilitate the generation of discrete, extraction-friendly outputs. Additionally, refining extraction algorithms to improve robustness and adaptability can further reduce the rate of invalid responses.


\section*{Ethics Statement}
This paper introduces the \textsc{SelectLLM} algorithm, a novel approach designed to leverage the diverse capabilities of various LLMs. While \textsc{SelectLLM} utilizes LLMs, it is crucial to recognize that, independent of this study, LLMs inherently present risks. These models may generate outputs that, despite being plausible, are factually inaccurate or nonsensical. Such \textit{hallucinations} can lead to misguided decision-making and the propagation of biases, particularly in high-stakes contexts where accuracy is paramount. In the absence of appropriate safeguards, the broad deployment of LLMs could exacerbate these issues. Thus, it is imperative to develop mechanisms that mitigate the risks of hallucinations to ensure the responsible and effective application of these models.




\bibliography{custom}

\appendix


\section{Prompting Templates and Answer Extraction}
\label{sec:pmtsamp}

Considering diverse LLMs and benchmarks adds challenges to prompting, as no single uniform prompting approach fits all LLMs \citep{sclar2023quantifying}. Based on insights from recent work on the appropriate usage of prompts \citep{sahoo2024systematic} and our own experiments, we make the following observations about prompting trends: 

\begin{enumerate}
    \item For non-chat LLMs, few-shot chain-of-thought (COT) prompting \citep{wei2022chain} works better than zero-shot \citep{kojima2022large} across both datasets. Therefore, we use five few-shot random examples obtained from the development set. The few-shot prompting results in over 95\% \textit{viable} answers (except for the {\tt llama2-7b-lm} LLM, which has a viability score of 83\%). An answer is considered \textit{viable} if it is represented by a single numeric/alphabetic string that can be extracted from the generated solution to compare with the reference answer. Viability is estimated using an automated regular expressions based-script and verified through manual inspection. 
    \item For chat LLMs, few-shot COT distracts the generator, leading to unexpected outputs, so zero-shot COT works best. To ensure correctness, we utilize different models' chat templates from HuggingFace.\footnote{\url{https://huggingface.co/docs/transformers/en/chat_templating}} The viability of answer extraction for chat models is approximately 92\%. Examples of zero-shot and few-shot prompting are presented in Appendix Figure \ref{fig:samprmpt}. 
\end{enumerate}
The limitation of answer extraction from 92\% to 95\% (except for {\tt llama2-7b-lm}) can be attributed to two factors: (i) limitations in the LLMs' response generation, where outputs are not structured in an extractable format, and (ii) limitations in the extraction algorithm, which fails to accurately parse the generated text. For more details, see the limitations section.

\begin{table*}[!t]
\centering
\resizebox{\linewidth}{!}{%
\begin{minipage}{0.35\textwidth}
  \centering
  \begin{tabular}{l|r|r}
    \hline \hline
    \textbf{Split}  & \textbf{GSM8K} & \textbf{MMLU} \\ \hline \hline
    Train & 6,816 & 13,757 \\ \hline
    Validation & 359 & 285 \\ \hline
    Test & 1,319 & 1,530 \\ \hline
    \hline 
  \end{tabular}
  \caption{Dataset statistics for \texttt{GSM8K} and \texttt{MMLU} benchmarks. For the \texttt{MMLU}, the officially released training and test splits have been swapped to align with the distribution of the \texttt{GSM8K}.
}
  \label{tab:data_stats}
\end{minipage}%
\begin{minipage}{0.70\textwidth}
  \centering
   \begin{tabular}{l|c|c|c}
    \hline \hline
    \textbf{LLMs}  & \textbf{Chat?} & \textbf{Spec?} & \textbf{\#Parameters}\\ \hline \hline
    \texttt{llama2-7b} & $\times$ & $\times$ & 7B \\ \hline
    \texttt{llama2-13b-chat} & \checkmark & $\times$ &  13B \\ \hline
    \texttt{mistral-7b} &  $\times$ & $\times$ & 7B\\ \hline
    \texttt{mistral-7b-it} & \checkmark & $\times$ & 7B \\ \hline
    \texttt{gemma-7b} & $\times$ & $\times$ & 7B\\ \hline
    \texttt{gemma-7b-it} & \checkmark & $\times$ & 7B \\ \hline
    \texttt{metamath-7b} & $\times$ & \checkmark & 7B\\
    \hline \hline
  \end{tabular}
  \caption{List of diverse LLMs selected for this study. \\ Spec: Specialized LLM}
  \label{tab:div_llms}
\end{minipage}%
}
\end{table*}

\begin{table*}[!t]
\centering
\resizebox{0.80\linewidth}{!}{%
\begin{tabular}{l|l|c|c}
\hline 
\hline
\textbf{LLM} &  \textbf{Prompt Type} & \textbf{GSM8K (prompt/sec)} & \textbf{MMLU (prompt/sec)} \\ \hline
& \texttt{llama2-7b} & 4.21 & 2.30 \\
& \texttt{gemma-7b} & 7.10 & 3.00 \\
Few-shot COT & \texttt{mistral-7b} & 3.70 & 1.10 \\
& \texttt{metamath-7b} & 4.70 & 2.40 \\ \hline
& \texttt{gemma-7b-it} & 0.70 & 1.00 \\
Zero-shot COT & \texttt{llama2-13b-chat} & 1.80 & 4.80 \\
& \texttt{mistral-7b-it} & 3.70 & 1.80 \\ \hline
\hline
\end{tabular}%
}
\caption{Runtime statistics on various LLMs over M generations for each input query. The timings are recorded using a single A100 GPU. `sec' denotes seconds, and COT denotes Chain-of-thought. For few-shot COT, we have considered 5 random examples from the development set.}
\label{tab:llm_lat}
\end{table*}

\begin{figure*}[!htb]
    \centering
    \includegraphics[width=1\linewidth]{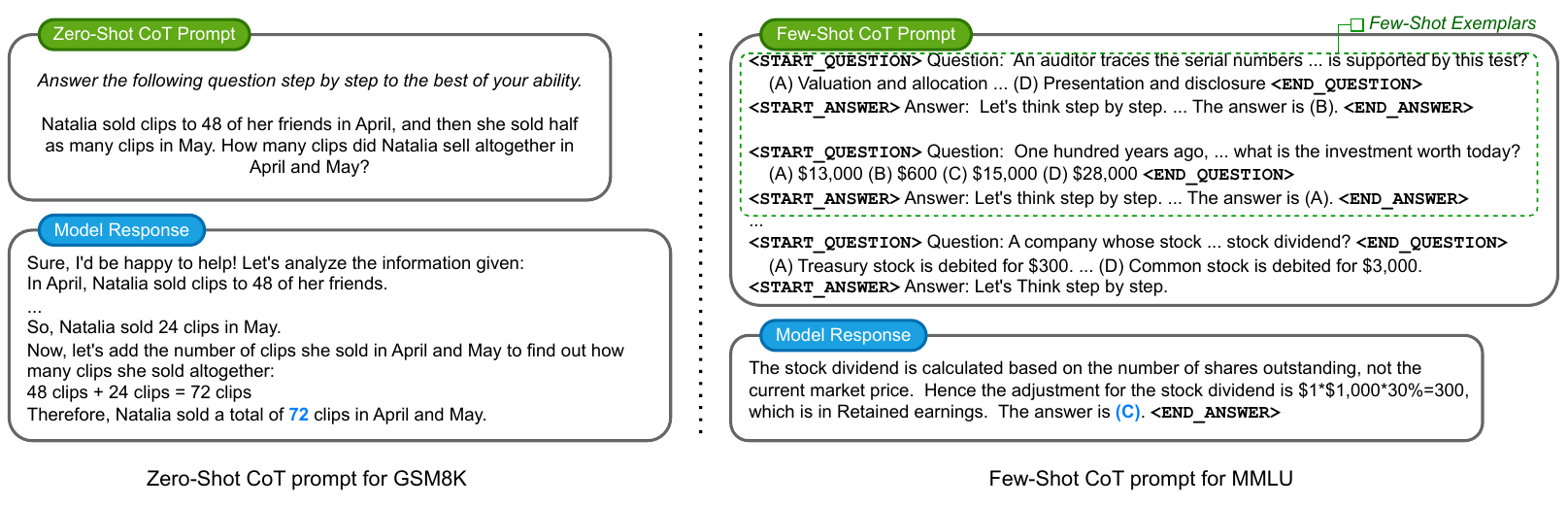}
    \caption{Sample zero-shot COT prompt for a chat (or instruction-tuned) LLM is shown for a GSM8K sample, and the few-shot COT prompt for non-chat LLMs is shown for an MMLU sample. Note that the type of prompting is associated with the LLM type rather than the dataset. Specifically, non-chat LLMs use few-shot COT, while chat models use zero-shot COT.}
    \label{fig:samprmpt}
\end{figure*}

The adapted prompting approaches used in our LLM queries are designed to instruct LLMs to specify that their final answers should be provided at the very end of each of their responses. We thus use a simple answer extraction policy of selecting the last mentioned numerical value (for {\tt GSM8K}) and multiple-choice option (for {\tt MMLU}) from the generated responses. Figure \ref{fig:samprmpt} in the Appendix shows a sample generation example. Responses failing to include any final answer are considered non-viable or invalid (\texttt{`INVALID'}) and counted as incorrect responses. For {\tt MMLU}, we evaluate the extracted options directly against the annotated correct answers (\texttt{`A'}, \texttt{`B'}, \texttt{`C'}, and \texttt{`D'}) from the dataset. For {\tt GSM8K}, questions where the absolute difference between the ground truth and predicted numerical answers is less than $\epsilon=0.1$ are evaluated as solved correctly. This threshold was set to accommodate instances where model-generated real-valued answers differ slightly from the expected answers, e.g., due to rounding errors.

\begin{figure*}[!t]
    \centering
    \begin{subfigure}[t]{0.5\textwidth}
        \centering
        \includegraphics[height=1.9in]{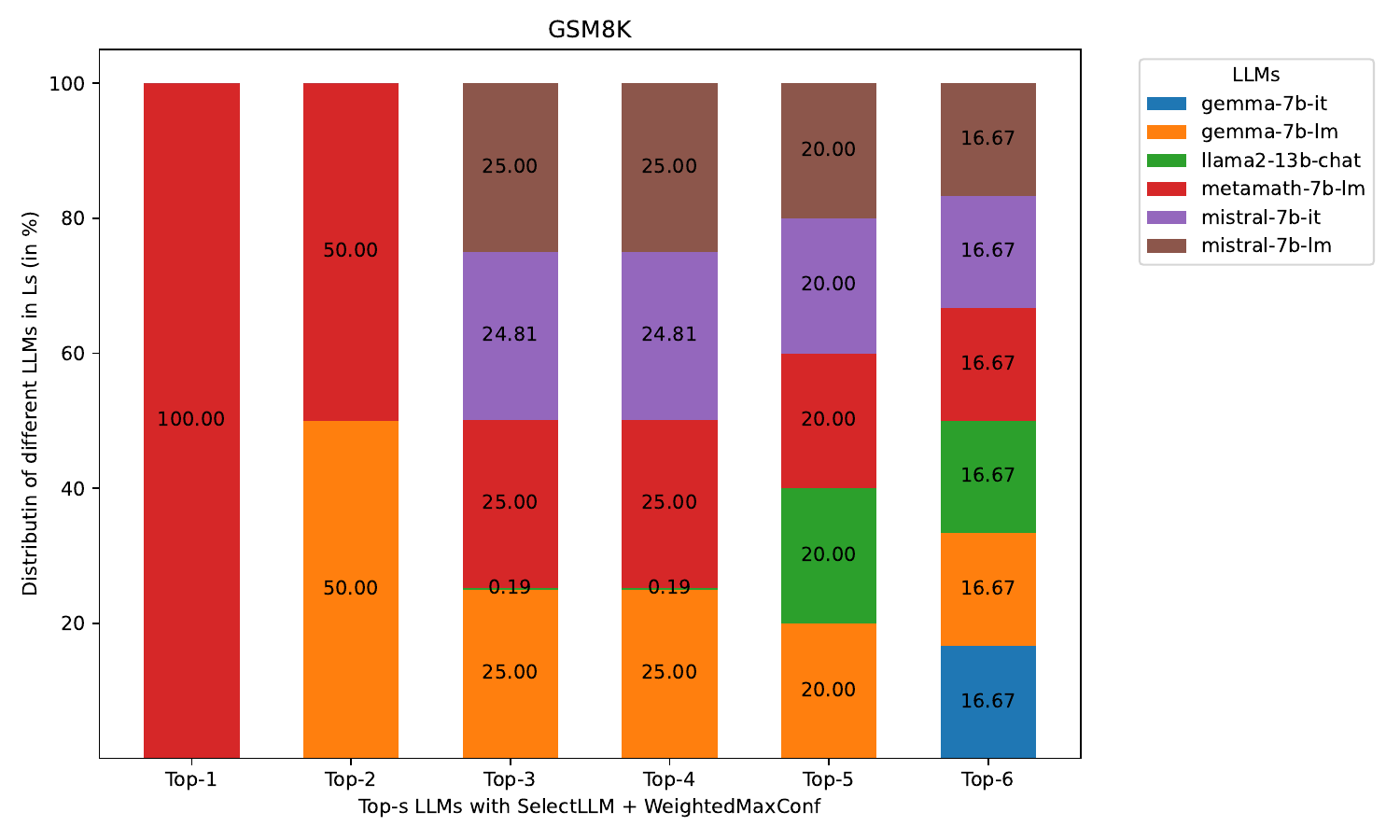}
    \end{subfigure}%
    \begin{subfigure}[t]{0.5\textwidth}
        \centering
        \includegraphics[height=1.9in]{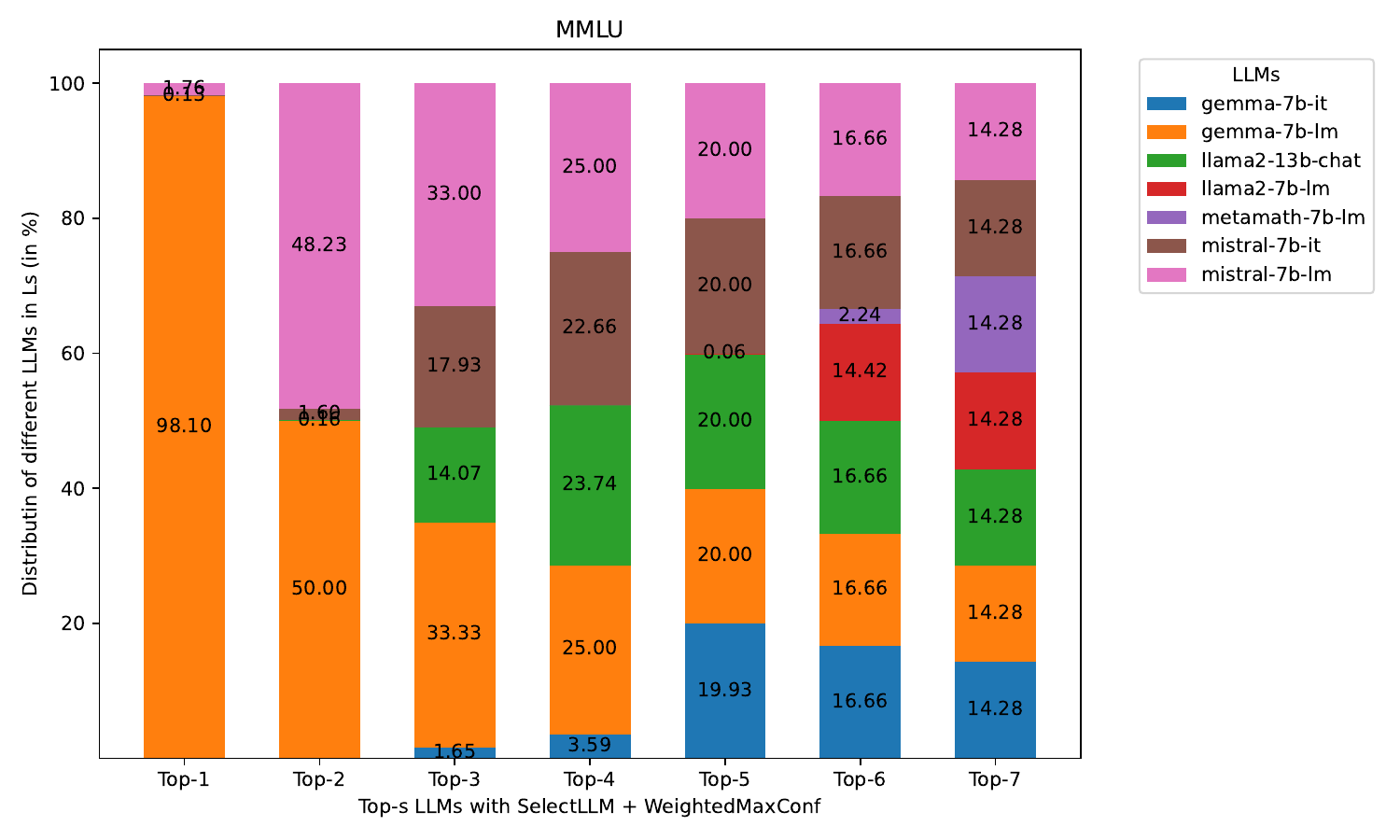}
    \end{subfigure}
    \caption{The distribution of different LLMs in the predicted subset of LLMs with \textsc{SelectLLM} algorithm for both {\tt GSM8K} (top) and {\tt MMLU} (bottom) datasets.}
    \label{fig:subsetllm_dist}
\end{figure*}

\begin{figure*}[!t]
    \centering
    \begin{subfigure}[t]{0.5\textwidth}
        \centering
        \includegraphics[height=1.7in]{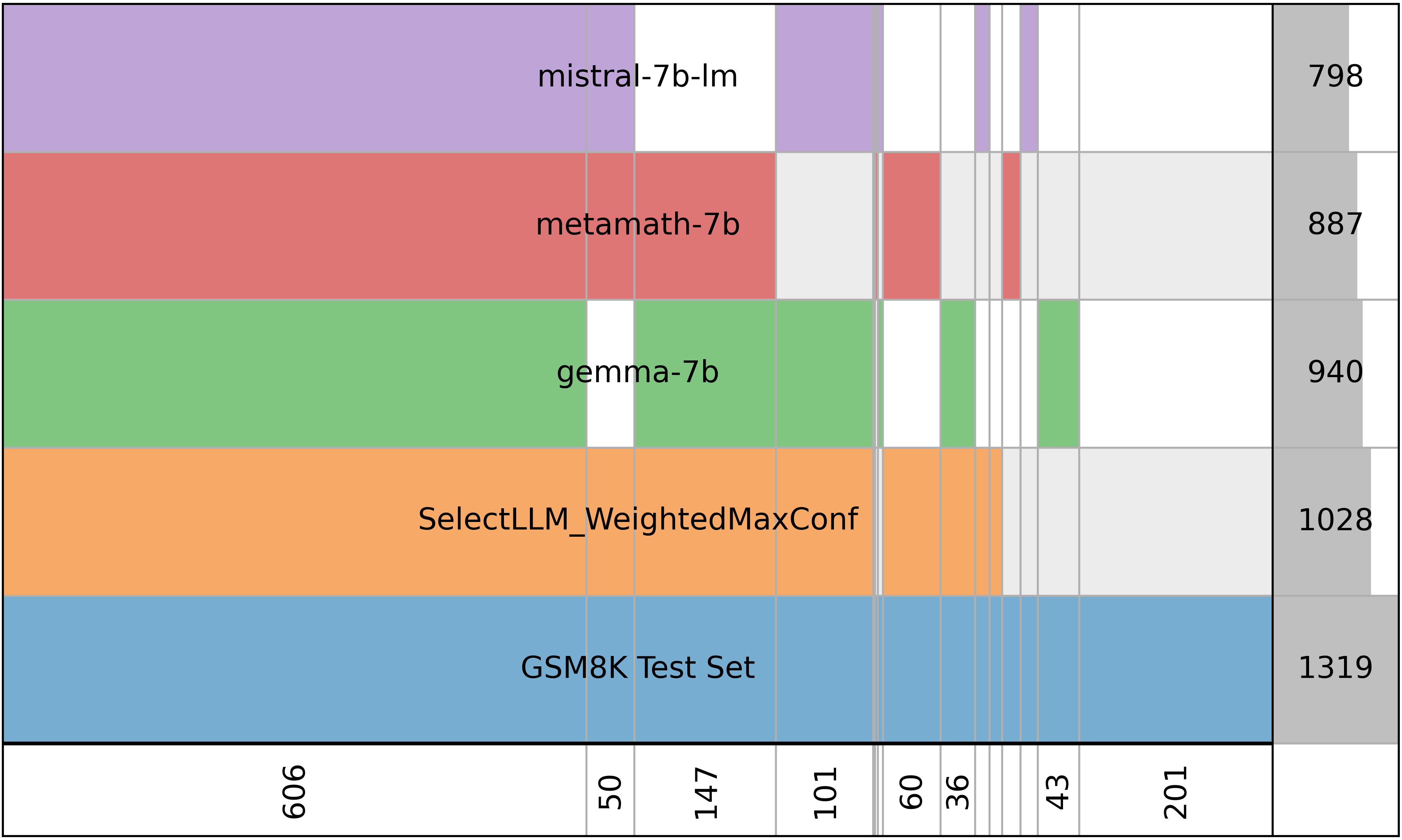}
    \end{subfigure}%
    ~ 
    \begin{subfigure}[t]{0.5\textwidth}
        \centering
        \includegraphics[height=1.7in]{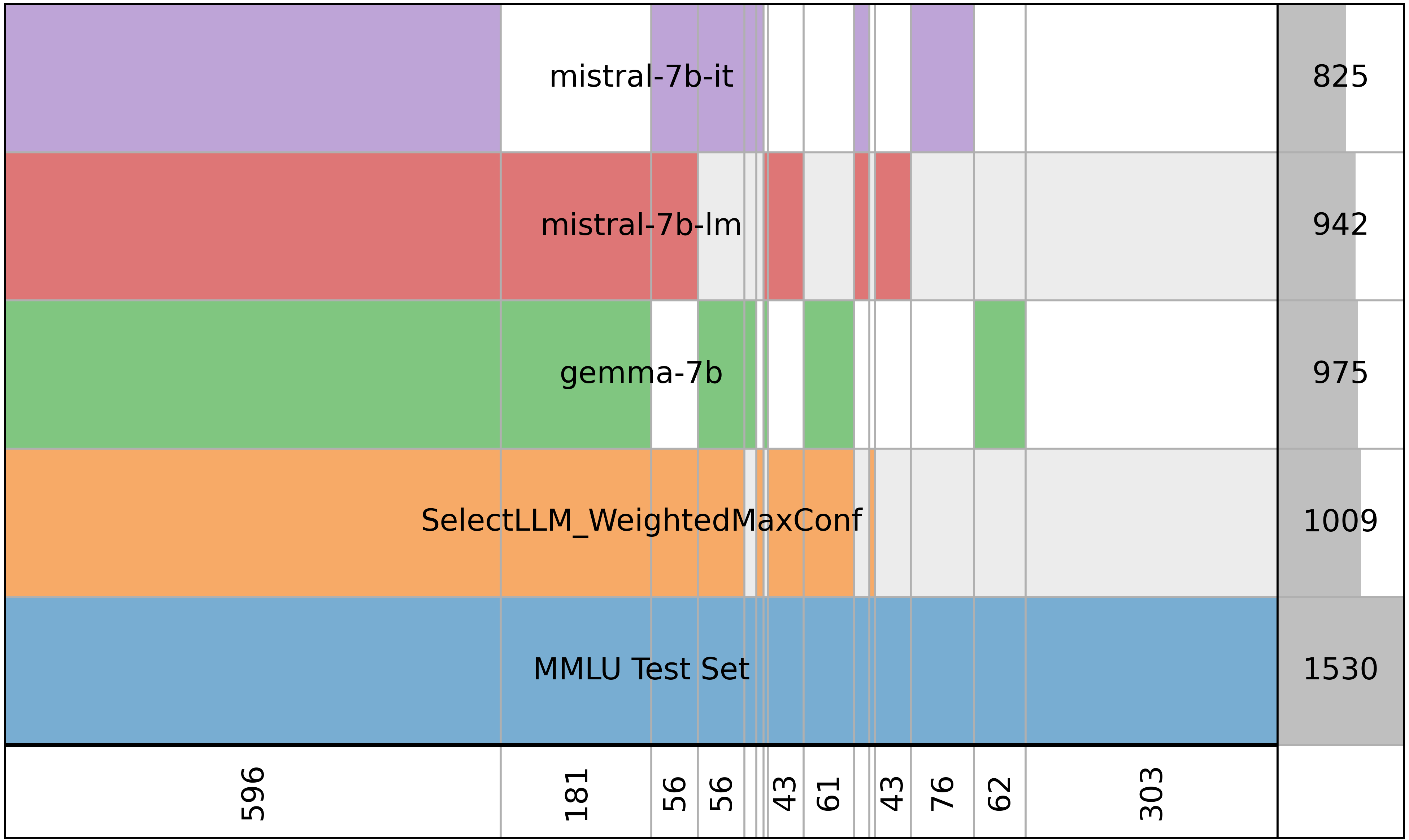}
    \end{subfigure}
    \caption{Distribution of the top-3 individual LLMs and proposed model responses to input queries for the test split of the {\tt GSM8K} (left) and {\tt MMLU} (right) datasets. The count in the rightmost column of each figure indicates the number of queries whose responses have been correctly answered by each LLM or the proposed model. The counts at the bottom denote the number of queries that have correct answers from one or more LLMs or the proposed model.}
    \label{fig:ques_dist}
\end{figure*}
  
\section{Data Statistics and Selected LLMs}
The statistics of the dataset and the list of considered diverse LLMs are presented in Table \ref{tab:data_stats} and \ref{tab:div_llms}, respectively.

\section{LLM Latency Estimation}
\label{sec:lat}
Runtime statistics of various LLMs are presented in Table \ref{tab:llm_lat}.

\section{Sample Prompts}
\label{sec:samprmpt}
Prompting examples of few-shot COT and zero-shot COT are illustrated in Figure \ref{fig:samprmpt}.

\begin{table}[!t]
\centering
\resizebox{\linewidth}{!}{%
  \begin{tabular}{l|r|r}
    \hline \hline
    \textbf{Model/Setup} & \textbf{GSM8K} & \textbf{MMLU} \\ \hline \hline
    BERT (base-uncased) & 0.63 & 0.59 \\ \hline
    BERT (base-uncased) + Weighted-label penalty & 0.64 & 0.62 \\ \hline
    RoBERTa (base) & 0.67 & 0.61 \\ \hline
    BRoBERTa (base) +  Weighted-label penalty & \textbf{0.71} & \textbf{0.68} \\ \hline
    T5 (base) & 0.66 & 0.60 \\ \hline
    T5 (base) + Weighted-label penalty & 0.69 & 0.65 \\ \hline
    \hline 
  \end{tabular}%
}
\caption{The F1 score for the MLC classifier is evaluated with different pre-trained language models. Additionally, a weighted-label penalty is applied to handle label imbalance, inspired by \cite{zhang2020towards}. This evaluation is conducted across both GSM8K and MMLU datasets.}
\label{tab:mlp_results}
\end{table}

\newpage 
\section{Distribution of Different LLMs in Top-$s$ Subset Selected with \textsc{SelectLLM}}
The details are shown in Figure \ref{fig:subsetllm_dist}.

\section{Query Response Distribution with Different Models} The details are presented in Figure \ref{fig:ques_dist}.

\section{Implementation Details}
\paragraph{Querying LLMs} We use the vLLM\footnote{https://github.com/vllm-project/vllm} package to query LLMs. All models were queried with a temperature of $0.8$ and a max token length of $2000$. Each question prompt was queried K times with different initialization seeds. We used a single NVIDIA A100 GPU for all runs. Querying each dataset once took approximately 1-2 hours.

\paragraph{MLC Training} We use the HuggingFace\footnote{https://huggingface.co/} library for loading and tuning all pre-trained Transformer encoders in our experiments. Each model was trained for $10$ epochs, with an initial learning rate of $1e$-$6$, a warmup ratio of $0.1$, and class-balanced CrossEntropy loss. The training checkpoint with the lowest validation loss was selected for inference. In this study, we have considered M=10 inspired by \citet{li2024more}.

\section{Performance of Individual LLMs}
Table \ref{tab:llm_results} presents the individual LLM performance for both the \texttt{GSM8K} and \texttt{MMLU} datasets. There is no clear best-performing model for both datasets. For instance, the \texttt{gemma-7b-lm} model performs the best in terms of accuracy for \texttt{GSM8K}, but its latency cost is also high. Similar trends are observed for the \texttt{MMLU} dataset. The performance of individual LLMs should not be directly compared with the \textsc{SelectLLM} model. Selecting the best LLM with high accuracy from a large pool requires running inferences across all LLMs to identify the top performer, resulting in the cumulative latency of all models, rather than just the selected one. On the other hand, randomly selecting an LLM may not necessarily choose the best performer. In this context, comparing \textsc{SelectLLM} with ensemble-based baselines is more appropriate, which is done in the main results (Table 2) where \textsc{SelectLLM} outperforms all baseline models, such as strange LLM-Blender baseline, in terms of both latency and accuracy.

\begin{table}[!t]
\centering
\resizebox{0.4\textwidth}{!}{%
\begin{tabular}{@{}l|cc|cc@{}}
\hline \hline
\multirow{2}{*}{\textbf{Models / Setups}} & \multicolumn{2}{c|}{\textbf{GSM8K}} & \multicolumn{2}{c}{\textbf{MMLU}} \\ \cline{2-5} 
 & \textbf{Acc $\uparrow$} & \textbf{Lat $\downarrow$} & \textbf{Acc $\uparrow$} & \textbf{Lat $\downarrow$} \\ \hline
\texttt{gemma-7b-lm} & 71.27 & 7.10 & 63.73 & 3.00 \\  
\texttt{mistral-7b-lm} & 60.50 & 3.70 & 61.57 & 1.80 \\ 
\texttt{metamath-7b-lm} & 67.25 & 4.70 & 41.76 & 2.40 \\  
\texttt{llama2-7b-lm} & -- & -- & 48.10 & 2.30 \\ 
\texttt{llama2-13b-chat} & 49.20 & 1.80 & 52.94 & 4.80 \\ 
\texttt{mistral-7b-it} & 56.71 & 1.00 & 53.92 & 1.10 \\  
\texttt{gemma-7b-it} & 42.23 & 0.70 & 50.72 & 1.00 \\ \hline \hline
\end{tabular}%
}
\caption{Performance and latency scores for different LLMs on {\tt GSM8K} \citep{cobbe2021training} and {\tt MMLU} \citep{hendryckstest2021} \textit{test} sets. {\tt Acc}: accuracy in percentage (\%); {\tt Lat}: latency in seconds; `--': values are not available.}
\label{tab:llm_results}
\end{table}

\section{Ablation Studies}
\begin{figure*}[!htb]
    \centering
    \begin{subfigure}[t]{0.5\textwidth}
        \centering
        \includegraphics[height=2in]{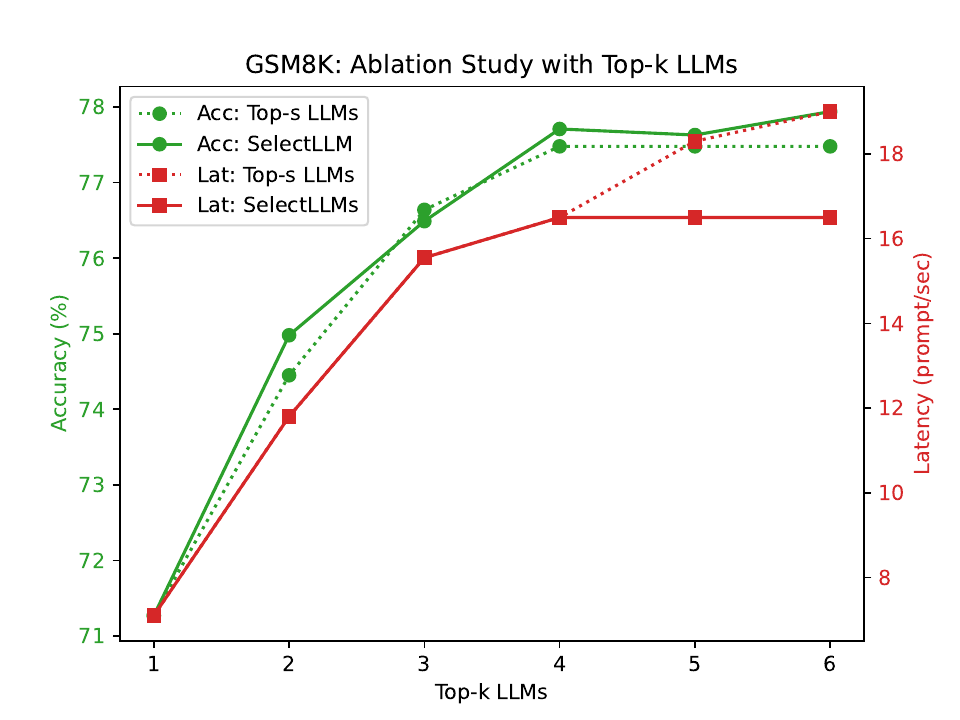}
    \end{subfigure}%
    ~ 
    \begin{subfigure}[t]{0.5\textwidth}
        \centering
        \includegraphics[height=2in]{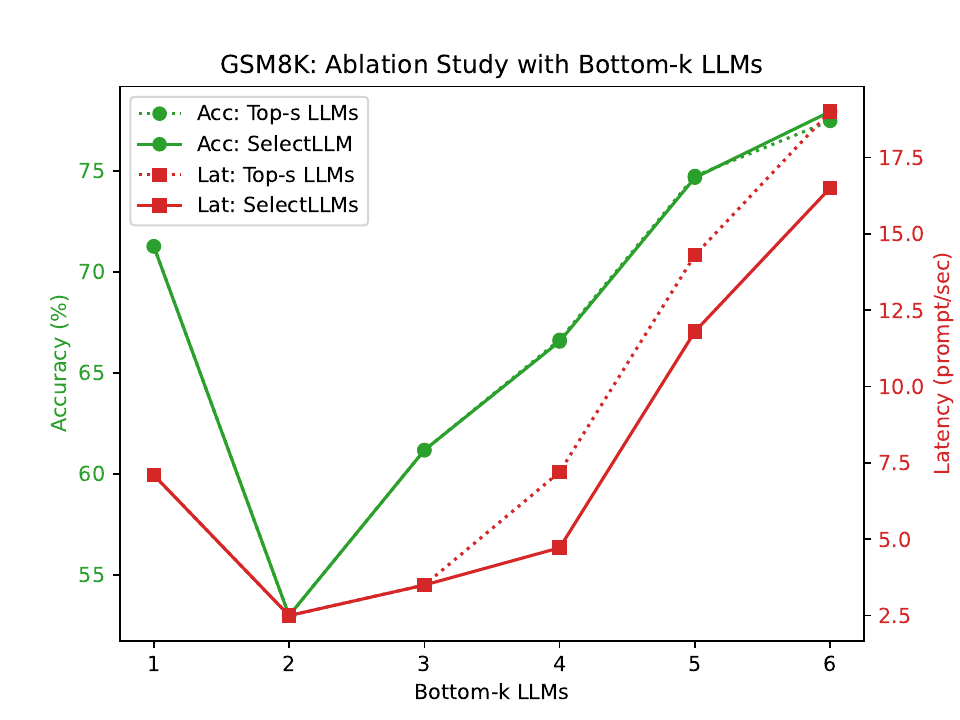}
    \end{subfigure}
    \caption{Ablation studies with top-$k$ and bottom-$k$ LLMs for {\tt GSM8K} dataset (left). Similar observations are made on the {\tt MMLU} dataset (right). For each LLM set, we have considered the values with an optimal $s$ for which the top-$s$ value is the highest.}
    \label{fig:ablation_study}
\end{figure*}

The details are presented in Figure \ref{fig:ablation_study}.

\section{Importance of Domain: A Case Study with MMLU}
\begin{figure}[t!]
    \centering
    \includegraphics[width=\columnwidth]{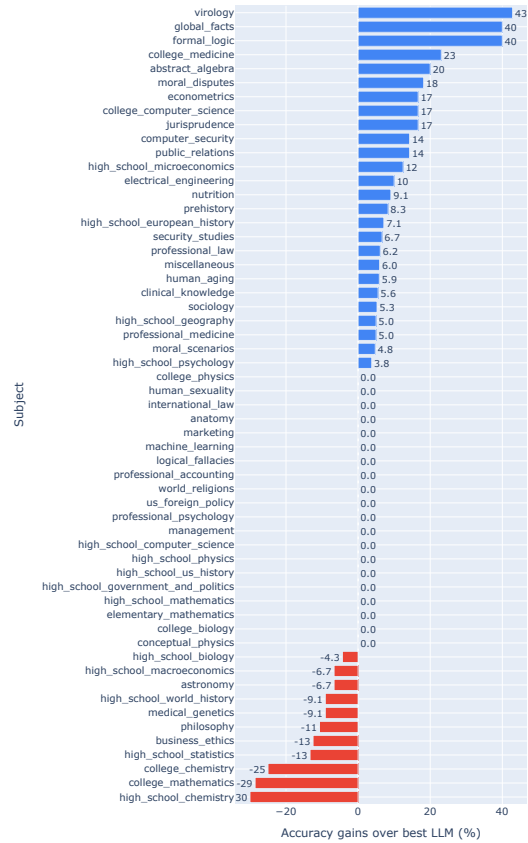}
    \caption{Subject-wise relative accuracy gain by {\sc Select}LLM with {\sc WeightedMaxConf} policy over the performance of the best-performing individual LLM (\texttt{gemma-7b-lm}).}
    \label{fig:mmlu_subject_analysis}
\end{figure}

The {\tt MMLU} dataset comprises 57 subjects. In this analysis, we evaluate the performance of the proposed \textsc{SelectLLM} algorithm using the {\sc WeightedMaxConf} policy, employing the best-performing individual LLM ({\tt gemma-7b-lm}) on a subject-wise basis. Figure \ref{fig:mmlu_subject_analysis} illustrates that while the proposed model's performance may be subpar for a few subjects, it demonstrates high performance for a significant portion of the subjects, indicating the effectiveness of the proposed model in general.

\section{SHAP Analysis}
\label{sec:shap-appendix}

In this section, we provide the details of the SHAP-values-based qualitative analysis discussed in Section \ref{sec:intpgap}.

\paragraph{Extracting Shapley Values} We first extract token-level Shapley values for the input question text to the fine-tuned MLC classifiers for each LLM using the \texttt{PartitionExplainer} module of the SHAP library \footnote{\href{https://shap.readthedocs.io/en/latest/}{\texttt{https://shap.readthedocs.io/en/latest/}}}. The token-level SHAP values are then summed according to word boundaries to get word-level SHAP values. This is a valid operation as SHAP values are additive in nature. 

\paragraph{Synset Mapping} In order to interpret the impact of different semantic word categories on the predicted solvability of questions, we map all viable words to their corresponding synsets using \texttt{NLTK}'s \texttt{WordNet} module \footnote{\href{https://www.nltk.org/}{\texttt{https://www.nltk.org/}}}. In this process, all instances of a word are recorded as instances of its respective synset. Next, we merge entries for synsets that represent less than 0.3\% (decided by observation) of the total number of tokens with their hypernyms. This helps reduce sparsity in the synsets to review. The mean SHAP values for each synset are then calculated across the test set of the datasets.

\paragraph{Synset Grouping} Based on the top 15 most detrimental synsets for each LLM in each of the four sets (i.e., having the lowest non-positive average SHAP values), we manually identify groups of related and recurring synsets. The groups for {\tt GSM8K}, as presented in Table \ref{tab:shap-gsm8k}, are as follows:
\begin{itemize}
    \item \textbf{\texttt{frequency terms}}: \\{\tt Synset('twice.r.01')}, {\tt Synset('thrice.r.01')}, {\tt Synset('times.n.01')}, etc.
    \item \textbf{\texttt{time duration units}}: \\{\tt Synset('hours.n.01')}, {\tt Synset('day.n.01')}, {\tt Synset('week.n.01')}, {\tt Synset('time\_unit.n.01')}, {\tt Synset('time\_period.n.01')}, etc.
    \item \textbf{\texttt{age units}}: Though similar to \texttt{time-units}, these synsets correspond specifically to instances where the age of an entity is being described. These include: {\tt Synset('age.n.01')}, {\tt Synset('year.n.01')}, {\tt Synset('time\_of\_life.n.01')}, etc.
    \item \textbf{\texttt{small numbers:}} Though some small integers and numbers posses their own individual synsets, we identify and filter all instances of small numbers ($\leq$100) (in numerical and word form) before synset matching and group them as \texttt{small numbers}. 
    \item \textbf{\texttt{quantifiers}}: \\ {\tt Synset('every.s.01')}, {\tt Synset('many.a.01')}, {\tt Synset('more.a.01')}, {\tt Synset('less.a.01')}, etc.
    \item \textbf{\texttt{ordinals}}:\\ {\tt Synset('first.a.01')}, {\tt Synset('second.s.01')}, {\tt Synset('third.s.01')}, etc.
    \item \textbf{\texttt{fractional values}}:\\ {\tt Synset('common\_fraction.n.01')}.
    \item \textbf{\texttt{rates and ratios}}:\\ {\tt Synset('rate.n.01')}, {\tt Synset('ratio.n.01')}, {\tt Synset('proportion.n.01')}, {\tt Synset('magnitude\_relation.n.01')}, etc.
    \item \textbf{\texttt{named entities}}: As \texttt{WordNet} does not record proper-nouns or named-entities like persons, organizations, locations, dates, and times, we identify such words before synset matching and designate them a separate group.
    \item \textbf{\texttt{other units}}: \\{\tt Synset('monetary\_unit.n.01')}, {\tt Synset('linear\_unit.n.01')}, {\tt Synset('work\_unit.n.01')}, {\tt Synset('area\_unit.n.01')}, {\tt Synset('definite\_quantity.n.01')}, etc.
    \item \textbf{\texttt{groups}}: \\{\tt Synset('family.n.01')}, {\tt Synset('team.n.01')}, {\tt Synset('unit.n.03')}, etc.
\end{itemize}

\end{document}